%% file: main.tex
\documentclass[11pt]{article}

\usepackage[preprint]{acl}

\usepackage{times}
\usepackage{latexsym}

\usepackage[T1]{fontenc}

\usepackage[utf8]{inputenc}

\usepackage{microtype}

\usepackage{inconsolata}
\usepackage{graphicx}
\usepackage{multirow}
\usepackage{url}
\usepackage{kotex}
\usepackage{float}
\usepackage{booktabs}
\usepackage{tabularx}
\usepackage{makecell}

\usepackage{listings}
\usepackage[most]{tcolorbox}

\usepackage{graphicx}
\usepackage{booktabs}

\usepackage{hyperref}
\usepackage{cleveref}
\usepackage{amsfonts}
\usepackage{pifont}
\usepackage{xcolor}

\usepackage{array}
\usepackage[table]{xcolor}

\definecolor{CIColor}{RGB}{234,242,252}
\definecolor{CVColor}{RGB}{235,247,239}
\newcommand{\scoredelta}[2]{%
#1\,{\fontsize{4.2}{4.6}\selectfont $\pm$#2}%
}

\title{Revealing Multi-View Hallucination in Large Vision-Language Models}

\author{
 \textbf{Wooje Park\textsuperscript{1*}},
 \textbf{Insu Lee\textsuperscript{1*}},
 \textbf{Soohyun Kim\textsuperscript{1}},
 \textbf{Jaeyun Jang\textsuperscript{1}},
\\
 \textbf{Minyoung Noh\textsuperscript{1}},
 \textbf{Kyuhong Shim\textsuperscript{2}},
 \textbf{Byonghyo Shim\textsuperscript{1}}
\\
 \textsuperscript{1}Seoul National University,
 \textsuperscript{2}Sungkyunkwan University
\\
{\small
  \texttt{\{wjpark, islee, soohyunkim, jyjang, mynoh, bshim\}@islab.snu.ac.kr}, 
  \texttt{khshim@skku.edu} }
  \\
}

\begin{document}
\maketitle

\renewcommand{\thefootnote}{}
\footnotetext{
\textsuperscript{$*$} Co-first authors.
}

\renewcommand{\thefootnote}{\arabic{footnote}} 
\input{sections/0_abstract}
\input{sections/1_introduction}
\input{sections/3_benchmark}

\input{sections/4_method}

\input{sections/5_experiment}

\section*{Limitations}
Although MVH-Bench provides a systematic benchmark for evaluating multi-view hallucination, it currently focuses on paired-view settings and does not cover more complex configurations involving a larger number of viewpoints. 
While RSCD performs consistently across the five LVLMs evaluated in this work, its effectiveness on other LVLMs remains to be further investigated.
Extending the benchmark to more diverse multi-view configurations and evaluating RSCD across a broader range of LVLMs would therefore be a promising direction for future work.

\section*{Acknowledgments}
This work was supported by the Institute of Information \& Communications Technology Planning \& Evaluation (IITP) under the Graduate School of Artificial Intelligence Semiconductor (IITP-2025-RS-2023-00256081), and by the National Research Foundation of Korea (NRF) (2022M3C1A3099336), both funded by the Korea government (MSIT).
The authors employed ChatGPT for language editing, including proofreading and stylistic refinement.
All AI-assisted revisions were subsequently reviewed and validated by the authors.

\bibliography{custom}

\clearpage 
\appendix
\input{sections/X_suppl}

\end{document}

%% file: sections/0_abstract.tex
\begin{abstract}
Large vision-language models (LVLMs) are increasingly being applied to multi-view image inputs captured from diverse viewpoints. 
Despite this growing use, current LVLMs often generate incorrect responses due to visual interference from non-target instances or viewpoints, a phenomenon we term multi-view hallucination (MVH).
To systematically analyze this problem, we construct MVH-Bench, a benchmark comprising 4.8k question-answer pairs targeting two types of hallucination: \textit{cross-instance} and \textit{cross-view}.
Empirical results show that MVH is prevalent across recent LVLMs.
To address this issue, we propose Reference Shift Contrastive Decoding (RSCD), a training-free decoding technique that suppresses visual interference by generating negative logits through attention masking.
Experiments on MVH-Bench with LLaVA-OneVision and Qwen2.5-VL demonstrate that RSCD improves performance by up to 25.7 and 94.8 points over existing hallucination mitigation methods, highlighting the effectiveness of RSCD.
\end{abstract}

%% file: sections/1_introduction.tex
\section{Introduction}
\input{figures/1_intro}
Large vision-language models (LVLMs) have shown remarkable advances in interpreting both visual and linguistic modalities and generating responses based on the given information~\cite{hurst2024gpt, liu2024llavanext,bordes2024introduction,li2025survey, vigsampler}.
Due to their capability to perceive complex visual scenes and describe them in natural language, LVLMs are widely employed as visual assistants in embodied AI, digital twins, and surveillance systems~\cite{majumdar2024openeqa, yuan2024surveillance, yang2024embodied, gholizadeh2025ai, shkim}.
A salient feature of these applications is that images captured from diverse viewpoints (i.e., multi-view images) are used as visual inputs to provide broader and more comprehensive information about a scene that a single-view image cannot offer.
Nevertheless, variations across different views of the same scene complicate the interpretation of rich visual information, making it difficult for LVLMs to produce precise responses~\cite{hong20233d, park2025bootstrap, lee2025towards}.

In recent years, several approaches to enhance inter-view image correspondence have been proposed.
One line of work aligns view-specific embeddings within a shared feature space, whereas another focuses on constructing a unified scene representation for comprehensive scene understanding~\cite{islam2023eqa, hong20233d, park2025bootstrap, lee2025towards}.
While these approaches enhance the information aggregation capabilities of LVLMs across multiple views, they fail to adequately judge which visual evidence corresponds to specific instances or viewpoints, resulting in significant perceptual errors.

Consider a scenario where LVLMs are queried about an instance in multi-view images (e.g., \textit{``What is the person wearing a white top doing?''}) (see Figure~\ref{fig:intro}).
Although the correct answer is \textit{``pointing at flowers''}, the model may confuse the queried instance with another instance, \textit{``the person wearing a black top''}, and incorrectly respond with \textit{``holding a watering can''}.
Similarly, when LVLMs are queried about a specific viewpoint, such as \textit{``In view 1, which direction is the watering can pointing?''}, the model may rely on another viewpoint (\textit{view 2}), and answer \textit{``pointing upward''} rather than the correct answer, \textit{``pointing to the left''}.

We refer to this phenomenon as multi-view hallucination (MVH), in which LVLMs generate incorrect responses due to visual interference across multiple viewpoints.
Building upon the examples described above, we categorize MVH into two types, depending on the cause of interference: \textit{cross-instance hallucination}, where the model’s responses include mixed details from other instances, and \textit{cross-view hallucination}, where the model references unintended viewpoints to produce responses.

To analyze MVH in LVLMs systematically, we design a new evaluation dataset dubbed MVH-Bench, comprising 4.8k question-answer pairs drawn from diverse multi-view scenarios.
For each multi-view image set, MVH-Bench includes two pairs of binary questions and one pair of multiple-choice questions generated by interchanging descriptions across instances and/or viewpoints.
Leveraging this paired-question structure, we design a set of dedicated evaluation metrics that measure the ability to provide answers grounded in sophisticated reasoning.

When we evaluate state-of-the-art (SOTA) closed and open-source LVLMs using the proposed MVH-Bench, we find that they are all confused by complex visual cues and thus struggle to generate correct answers.
For example, we observe that models tend to generate incorrect responses based on visual cues associated with certain words or partial phrases in the query.
To better investigate these phenomena, we compare the model’s responses before and after disrupting query context formation by blocking interactions between text tokens at each layer.
Interestingly, we observe under this condition that the model’s responses shift toward non-target visual cues.
This implies that information exchange between text tokens in the intermediate layers plays a key role in accurate visual grounding.

To alleviate visual grounding errors arising from insufficient query understanding, we propose a novel contrastive decoding technique referred to as Reference Shift Contrastive Decoding (RSCD).
The core idea of RSCD is to amplify the contrast between the original logits and those obtained with incomplete textual understanding, to suppress predictions grounded in non-target visual cues.
To do so, we first obtain negative logits by intentionally disrupting text-to-text attention pathways in the intermediate layers, and then adjust the original logits in the direction opposite to the difference.
Our experiments on MVH-Bench using LLaVA-OneVision~\cite{li2024llava} and Qwen2.5-VL~\cite{bai2025qwen2} show that RSCD consistently outperforms the recent decoding technique AvisC~\cite{woo2024dont} by 25.7 and 94.8 points in overall benchmark score, respectively.

\input{figures/2_benchmark}
\noindent In summary, our main contributions are as follows:
\begin{itemize}
    \item In this work, we characterize the MVH problem as an error caused by visual interference across multiple viewpoints and categorize it into two error types based on the source of the confusion.
    
    \item In order to scrutinize the MVH problem in detail, we design a new benchmark named MVH-Bench and propose dedicated accuracy metrics, viz., p-Acc and q-Acc, using which we evaluate 23 SOTA LVLMs.

    \item Through extensive experiments, we show that RSCD outperforms recent hallucination mitigation methods, improving MVH-Bench scores by up to 25.7 and 94.8 for LLaVA-OneVision and  Qwen2.5-VL, respectively.
\end{itemize}

%% file: figures/1_intro.tex
\begin{figure*}[t!]
\centering
\includegraphics[width=1.0\linewidth]{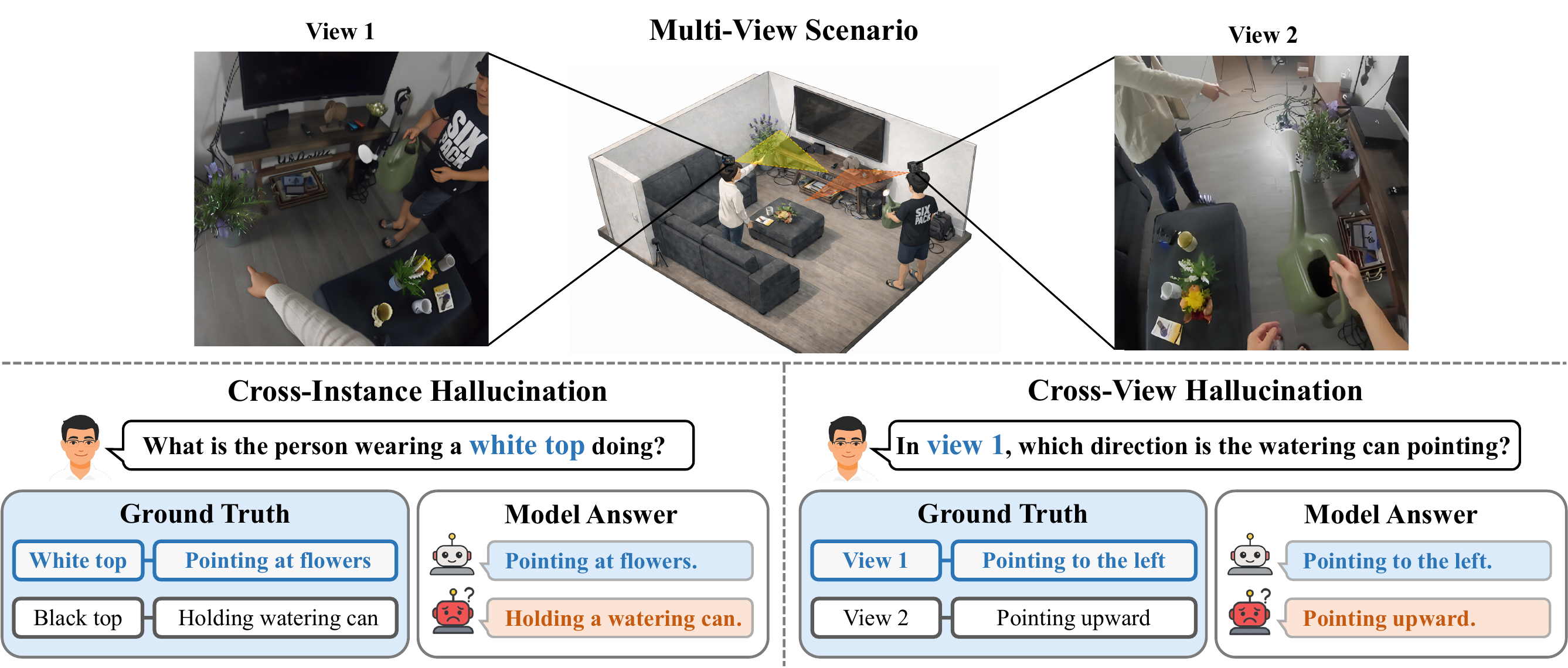}
\caption{Illustration of two types of multi-view hallucination in LVLMs, categorized by the source of the interference.
\emph{Cross-Instance Hallucination}: The model relies on information from another instance.
\emph{Cross-View Hallucination}: The model relies on information from another view.
}
\vspace{-0.3cm}
\label{fig:intro}
\end{figure*}

%% file: figures/2_benchmark.tex
\begin{figure*}[t!]         
\centering
\includegraphics[width=\textwidth]{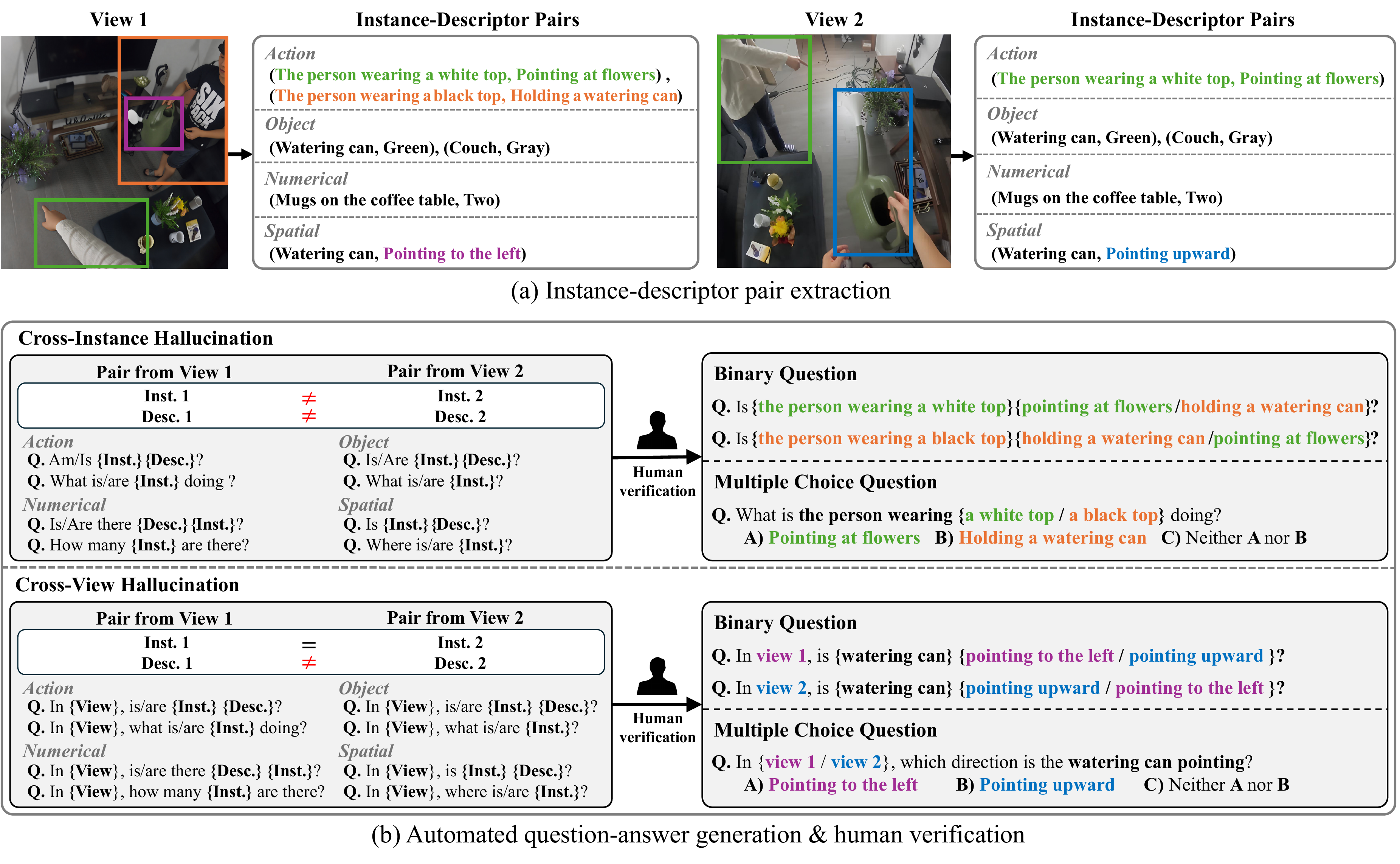}
\caption{
Overview of the MVH-Bench construction pipeline:
(a) instance-descriptor pair extraction, and
(b) automated question-answer generation followed by human verification.
}
\label{fig:benchmark}
\end{figure*}

%% file: sections/3_benchmark.tex
\section{Multi-View Hallucination Benchmark}
\subsection{Benchmark Design}

In order to construct MVH-Bench, we use Ego-Exo4D~\cite{10658224egoexo4d} and LEMMA~\cite{jia2020lemma}, both of which provide synchronized multi-view videos of diverse scenes involving multi-person interactions.
Following the procedure of~\citet{lee2025towards}, we extract image frames from Ego-Exo4D and use the frames provided by the authors for LEMMA.
We then pair images from diverse combinations of first-person and third-person views to cover a broad range of multi-view scenarios.
The dataset construction process consists of three steps: 1) Instance-descriptor pair extraction, 2) Automated question-answer generation, and 3) Human verification (see Figure~\ref{fig:benchmark} for an overview).

\subsubsection{Instance-Descriptor Pair Extraction}

MVH-Bench includes four subcategories that closely reflect real-world scenarios: \textit{action}, \textit{object}, \textit{numerical}, and \textit{spatial}~\cite{lee2025towards}.
For each subcategory, we extract subcategory-specific \textit{instance-descriptor} pairs $(I, D)$ from each view of every image pair using GPT-4o.
Here, an \textit{instance} uniquely identifies a specific person or object in an image, while a \textit{descriptor} describes the state or attribute of that instance.
For example, in the action subcategory, an instance may be \textit{``a person wearing a black shirt''} and the corresponding descriptor may be \textit{``holding a smartphone''}. 
Given a multi-view image pair, let 
$\mathcal{S}_{1}=\{(I_{i},D_{i})\}_{i=1}^{n_{1}}$ and 
$\mathcal{S}_{2}=\{(I_{j},D_{j})\}_{j=1}^{n_{2}}$ 
denote the sets of extracted instance-descriptor pairs from the first and second views, respectively. 
The prompts used to extract instance-descriptor pairs are provided in Appendix~\ref{appendix:prompt}.

\subsubsection{Automated Question-Answer Generation}

We design MVH-Bench using two types of questions, binary and multiple-choice, each serving a distinct purpose.
Using a binary question-answer (QA) format, we evaluate the model’s ability to correctly understand the relationships between instances and their corresponding descriptors or views.
With a multiple-choice QA format, we assess its ability to distinguish the correct descriptor from plausible distractors.
Using the instance-descriptor pairs from $\mathcal{S}_{1}$ and $\mathcal{S}_{2}$, we automatically generate questions for the two MVH categories: \textit{cross-instance} and \textit{cross-view}.
For clarity, we describe the process using a single image pair from the action subcategory.

\paragraph{Cross-Instance Hallucination.}

We observe that models often confuse the queried instance with other instances and incorrectly respond based on their descriptors.
To systematically diagnose this behavior, we evaluate the model's ability to correctly ground the queried instance among other instances.
Concretely, we sample $(I_i, D_i)$ from $\mathcal{S}_1$ and $(I_j, D_j)$ from $\mathcal{S}_2$ such that the instances and descriptors are mutually distinct (i.e., $I_i \neq I_j$ and $D_i \neq D_j$).
We then construct binary and multiple-choice QA pairs using these.
For the binary questions, we generate four QAs following the template:
\begin{quote}
    \textbf{Q:} \texttt{``Am/Is $I_x$ $D_y$?''}

    \textbf{A:} \texttt{``Yes''} if $x = y$, \texttt{``No''} otherwise.
\end{quote}
For the multiple-choice questions, we generate two QA pairs using the template:
\begin{quote}
    \textbf{Q:} \texttt{``What is $I_x$ doing?''}

    \hspace{1em} A) \texttt{``$D_x$''} \qquad\qquad B) \texttt{``$D_y$''}

    \hspace{1em} C) \texttt{``Neither $D_x$ nor $D_y$''}

    \textbf{A:} A) \texttt{``$D_x$''}
\end{quote}

\paragraph{Cross-View Hallucination.}

Another failure mode is that the model tends to focus on instances from the non-queried view.
To rigorously diagnose this phenomenon, we evaluate the model's ability to correctly ground instance-descriptor pairs in the appropriate view.
Specifically, in our benchmark, we deliberately design binary and multiple-choice QAs using pairs \((I_i, D_i)\) and \((I_j, D_j)\) with identical instances yet having distinct descriptors (i.e., \(I_i = I_j\) and \(D_i \neq D_j\)).
This design ensures that the same instance can be queried across different views.
For the binary case, we generate four QA pairs following the template:
\begin{quote}
    \textbf{Q:} \texttt{``In view $x$, is $I_x$ $D_y$?''}

    \textbf{A:} \texttt{``Yes''} if $x = y$, \texttt{``No''} otherwise.
\end{quote}
For the multiple-choice case, we generate two QA pairs using the template:
\begin{quote}
    \textbf{Q:} \texttt{``In view $x$, what is $I_x$ doing?''}
    
    \hspace{1em} A) \texttt{``$D_x$''} \qquad\qquad B) \texttt{``$D_y$''}

    \hspace{1em} C) \texttt{``Neither $D_x$ nor $D_y$''}

    \textbf{A:} A) \texttt{``$D_x$''}
\end{quote}
During evaluation, answer options in multiple-choice questions are randomly shuffled to make sure correct answers are distributed uniformly.

\subsubsection{Human Verification}
After automated QA generation, human annotators review every QA pair and remove those describing non-existent and/or incorrect content, as well as pairs that are trivially easy to solve.
The resulting MVH-Bench consists of 3,200 binary and 1,600 multiple-choice QAs, evenly distributed across hallucination types and subcategories.
The dataset is split into test and validation sets in a 9:1 ratio while preserving the original distribution.
The validation set is used for hyperparameter search and analysis, while the test set is reserved for the final evaluation.
Additional details of the human verification stage are provided in Appendix~\ref{appendix:human verification}.

\newcommand{\I}{\mathbb{I}}

\subsection{MVH-Bench Evaluation}
\label{sec:metric}
Utilizing the paired-question format in MVH-Bench, we propose a dedicated metric to assess whether a model correctly grounds its responses.
For a formal expression of the evaluation metrics, please refer to Appendix~\ref{appendix:eval metrics}.

\input{figures/3_method}

\subsubsection{Binary Question}
We evaluate the performance on binary questions using three accuracy metrics. 
The first is the mean accuracy (Acc) across all four QA pairs.
In our setup, even questions whose correct answer is ``No'' contain elements that are present in the multi-view images.
Consequently, it is highly likely that the model generates the answer ``Yes'' exclusively.
To assess the model’s ability to truly ground the queried instance, we measure pair accuracy (p-Acc), which evaluates if the model correctly answers both the Yes and No questions associated with each view.
Additionally, a model may appear to perform well by relying on a single view while failing to understand others.
To verify consistent understanding across all views, we further measure quadruplet accuracy (q-Acc).
In this metric, we consider a prediction correct only when all four binary questions are answered correctly.
To quantify the model's tendency toward biased responses, we also compute the yes-error ratio (YER), which is the fraction of Yes predictions among the incorrect cases.

\subsubsection{Multiple-Choice Question}
For the multiple-choice setting, we measure the mean accuracy (Acc) and pair accuracy (p-Acc).
In pair accuracy, to assess model consistency across views, we consider a prediction correct when both questions derived from each multi-view image are answered correctly.
In addition, to analyze the model’s error tendencies, we report the fraction of incorrect answers that choose the adversarial option B ($D_y$) over the neutral option C.
We refer to this as the adversarial error ratio (AER).
A high AER value indicates that the model tends to select the adversarial distractor.

Finally, we define the overall model performance, denoted as MVH Score, as the sum of all evaluation metrics except YER and AER.

%% file: figures/3_method.tex
\begin{figure*}[t!]
  \centering
  \includegraphics[width=\textwidth]{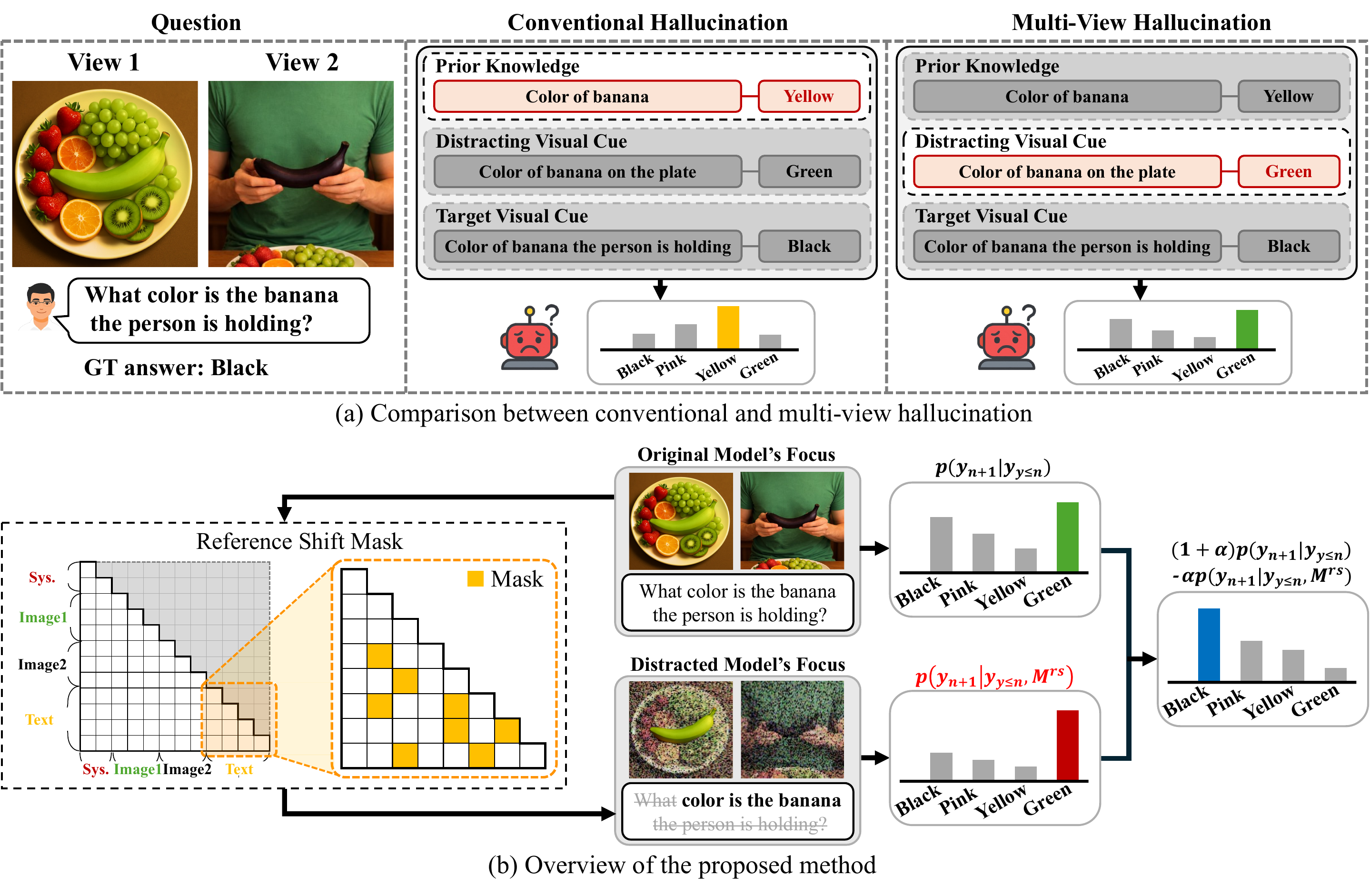}
    \vspace{-6pt}
  \caption{
(a) Comparison between conventional and multi-view hallucination by their underlying causes.
(b) Overview of the proposed RSCD, illustrating its core idea and underlying intuition.
  }
  \label{fig:method}
    \vspace{-6pt}
\end{figure*}

%% file: sections/4_method.tex
\section{Reference Shift Contrastive Decoding}

\input{tables/baseline}

\subsection{MVH Problem Statement}
Multi-view hallucination (MVH) is fundamentally distinct from conventional hallucination in that it arises when models ground their responses in visual information from non-target instances or viewpoints.
As illustrated in Figure~\ref{fig:method}(a), conventional hallucination occurs when the model relies on prior knowledge (e.g., \textit{the color of a banana is yellow}).
In contrast, MVH occurs when the model is distracted by spurious visual cues (e.g., \textit{the color of the banana on the plate is green}).
Therefore, an approach that simply makes the model attend more to \textit{all} visual cues would not work in most cases.
A better way to address MVH is to guide the model to attend to the \textit{correct} visual evidence.
We argue that understanding the full context of the query is essential for the model to identify the correct visual evidence.
When the model understands the query only partially, it may attend to visual content that is consistent with only part of the query.
To empirically examine this hypothesis, we analyze how query understanding emerges in LVLMs and also contributes to visual grounding.

\subsection{Identifying Layers for Query Understanding}
\label{3.2} 
In LVLMs, system tokens, image tokens, and text tokens are concatenated into a single sequence of length $T$ and processed by a Transformer decoder.
At each decoder layer $l \in \{0,\dots,L-1\}$, these tokens are updated through self-attention, with the attention matrix $A \in \mathbb{R}^{T \times T}$ computed as follows:
\begin{equation}
A = \mathrm{softmax}(QK^\top + M^c),
\end{equation}
where $Q$ and $K$ are the query and key matrices, respectively.
The causal mask $M^c$ is defined as
\begin{equation}
M^c_{i,j} =
\begin{cases}
-\infty, & j > i, \\
0, & \text{otherwise}.
\end{cases}
\end{equation}
The attention matrix $A$ determines how token representations are updated, with $A_{i,j}$ representing the amount of information flow from the $j$-th token to the $i$-th token~\cite{zhang2025cross-modal-information-flow}.
Since self-attention is the only mechanism for information exchange across token positions, query understanding likely emerges through interactions among text tokens within self-attention.
Under this view, blocking attention between text tokens may hinder the model’s ability to understand the query.
To examine this effect, we employ an additional mask $M^{\mathrm{t2t}}$ that blocks attention from text tokens to text tokens: 
\begin{equation}
M^{\mathrm{t2t}}_{i,j} =
\begin{cases}
-\infty, & i,j \in \mathcal{T}, \\
0, & \text{otherwise},
\end{cases}
\end{equation}
where $\mathcal{T}$ denotes the set of text-token indices.
We apply this mask over a sliding window of $w$ consecutive decoder layers to analyze how the effect of blocking text-to-text information flow varies across layer ranges.
Concretely, we perform a captioning task on image pairs, asking the model to describe either the first or the second image in each pair (see Appendix~\ref{appendix:captioning} for details).
To quantify the resulting response changes, we define \textit{reference accuracy}, which considers a response correct if the generated caption is grounded in the corresponding visual evidence.
As shown in Figure~\ref{fig:interaction-window}, reference accuracy drops when the mask is applied to middle layers (e.g., layers 13--20 in LLaVA-OneVision).
This suggests that blocking text-to-text attention in these layers may lead the model to reference the incorrect regions of the image.
Based on this observation, we identify this layer range, denoted by $\Lambda^{\star}$, as important for query understanding and visual grounding.
We provide additional analysis in Appendix~\ref{appendix:t2t_analysis} that further supports our hypothesis.

\input{tables/baseline_bias}

\input{tables/method}

\subsection{Contrastive Decoding via Reference Shift}
To mitigate MVH, we deliberately amplify the effects of incomplete query understanding and subtract the resulting bias from the original logits.
Specifically, for each layer $l \in \Lambda^{\star}$, we mask the top $\rho\%$ of attention entries in each row corresponding to a text token $i \in \mathcal{T}$.
Here, $\rho$ controls how strongly textual context formation is disrupted.
We define the resulting mask as the reference shift mask (RSM), denoted by $M^{rs}$:
\begin{equation}
M^{rs}_{i,j} =
\begin{cases}
-\infty, & i \in \mathcal{T} \ \text{and}\ j \in \mathcal{J}_{i}, \\[4pt]
0,        & \text{otherwise},
\end{cases}
\label{eq:toprho-drop}
\end{equation}
where $\mathcal{J}_{i} = \{\, j \mid A_{i,j} \in \mathrm{TopP}(A_{i,\mathcal{T}}, \rho) \,\}$ is the set of text-token indices that receive the top $\rho\%$ attention from token $i$ within $A_{i,\mathcal{T}}$.

A forward pass with $M^{rs}$ yields the negative logit 
$\operatorname{logit}(y_{n+1} \mid y_{\le n}, ~M^{rs} )$, where $y_{\le n}$ denotes the input sequence up to position $n$.
Let $\operatorname{logit}(y_{n+1} \mid y_{\le n})$ be the base logit from the unmodified model.
We then contrast the base and negative logit at each decoding step as follows:
\begin{align}
\operatorname{logit}^{rscd}
&= (1+\alpha)\,\operatorname{logit}(y_{n+1} \mid y_{\le n}) \notag\\
&\quad - \alpha\,\operatorname{logit}(y_{n+1} \mid y_{\le n}, ~M^{rs}),\end{align}
where $\alpha$ scales the amplification of the differences between the two logits.
Following previous work, we apply adaptive plausibility constraints during decoding to consider only high-probability tokens under the original distribution~\cite{li2023contrastive-decoding}.
The overall decoding process of RSCD is illustrated in Figure~\ref{fig:method}(b).

\input{figures/3_1_analysis}

%% file: tables/baseline.tex
\begin{table*}[tb]

\centering
\scriptsize

\setlength{\tabcolsep}{1.0pt}
\renewcommand{\arraystretch}{1.25}

\resizebox{\textwidth}{!}{%
\begin{tabular}{@{}l
!{\vrule width 0.9pt}
>{\centering\arraybackslash}p{0.82cm}
>{\centering\arraybackslash}p{0.82cm}
>{\centering\arraybackslash}p{0.82cm}
|>{\centering\arraybackslash}p{0.82cm}
>{\centering\arraybackslash}p{0.82cm}
!{\vrule width 0.5pt}
>{\centering\arraybackslash}p{1.3cm}
!{\vrule width 0.9pt}
>{\centering\arraybackslash}p{0.82cm}
>{\centering\arraybackslash}p{0.82cm}
>{\centering\arraybackslash}p{0.82cm}
|>{\centering\arraybackslash}p{0.82cm}
>{\centering\arraybackslash}p{0.82cm}
!{\vrule width 0.5pt}
>{\centering\arraybackslash}p{1.3cm}
!{\vrule width 0.9pt}
>{\centering\arraybackslash}p{1.30cm}
@{}}

\toprule

\multirow{3}{*}{\textbf{LVLMs}}
&
\multicolumn{6}{c!{\vrule width 0.9pt}}
{\textbf{Cross-Instance}}
&
\multicolumn{6}{c!{\vrule width 0.9pt}}
{\textbf{Cross-View}}
&
\multirow{3}{*}{\makecell{\textbf{MVH}\\\textbf{Score}}}
\\

\cmidrule(lr){2-7}
\cmidrule(lr){8-13}

&
\multicolumn{3}{c|}{\textbf{Binary}}
&
\multicolumn{2}{c!{\vrule width 0.5pt}}{\textbf{M.C.}}
&
\multirow{2}{*}{\textbf{Score}}
&
\multicolumn{3}{c|}{\textbf{Binary}}
&
\multicolumn{2}{c!{\vrule width 0.5pt}}{\textbf{M.C.}}
&
\multirow{2}{*}{\textbf{Score}}
&
\\

&
\textbf{Acc}
&
\textbf{p-Acc}
&
\textbf{q-Acc}
&
\textbf{Acc}
&
\textbf{p-Acc}
&
&
\textbf{Acc}
&
\textbf{p-Acc}
&
\textbf{q-Acc}
&
\textbf{Acc}
&
\textbf{p-Acc}
&
&
\\

\midrule

\multicolumn{14}{c}{\textbf{Closed-Source}} \\

\midrule

GPT-4o
& \textbf{73.66}
& \textbf{52.41}
& \textbf{32.04}
& 57.27
& 36.11
& \scoredelta{\textbf{251.49}}{0.32}
& 68.87
& 46.48
& 23.43
& 55.74
& 32.31
& \scoredelta{226.83}{0.16}
& \scoredelta{\underline{478.32}}{0.47} \\

GPT-4o mini
& 62.20
& 31.53
& 12.87
& \underline{66.99}
& \textbf{44.81}
& \scoredelta{\underline{218.40}}{0.77}
& 61.92
& 35.83
& 14.72
& 61.06
& 33.06
& \scoredelta{206.59}{0.72}
& \scoredelta{424.99}{0.67} \\

Gemini 2.5 Flash
& \underline{64.42}
& \underline{39.68}
& \underline{14.91}
& 60.83
& 38.15
& \scoredelta{217.99}{2.34}
& \textbf{71.00}
& \textbf{51.16}
& \textbf{28.06}
& 69.54
& 48.89
& \scoredelta{\textbf{268.65}}{2.47}
& \scoredelta{\textbf{486.64}}{2.90} \\

Gemini 2.0 Flash
& 60.72
& 29.58
& 9.63
& \textbf{68.15}
& \underline{44.72}
& \scoredelta{212.80}{3.18}
& \underline{69.91}
& 44.54
& 23.24
& \textbf{73.06}
& \textbf{53.80}
& \scoredelta{\underline{264.55}}{2.49}
& \scoredelta{477.35}{5.66} \\

Claude Sonnet 4.5 
& 61.44
& 36.94
& 13.61
& 60.00
& 36.02
& \scoredelta{208.01}{0.87}
& 69.26
& \underline{49.35}
& \underline{23.61}
& \underline{72.03}
& \underline{49.42}
& \scoredelta{263.67}{3.20}
& \scoredelta{471.68}{4.05} \\

\midrule

\multicolumn{14}{c}{\textbf{Open-Source}} \\

\midrule

InternVL3-14B
& \textbf{64.68}
& \underline{40.69}
& \textbf{18.33}
& \underline{68.47}
& \underline{47.59}
& \scoredelta{\textbf{239.76}}{3.70}
& 55.37
& 29.03
& 4.48
& 56.90
& 26.11
& \scoredelta{171.89}{6.53}
& \scoredelta{411.65}{9.67} \\

Gemma3-12B
& 55.76
& 18.52
& 3.89
& 59.21
& 33.24
& \scoredelta{170.62}{5.12}
& 58.35
& 24.91
& 5.09
& 58.89
& 30.37
& \scoredelta{177.61}{0.94}
& \scoredelta{348.23}{5.98} \\

Llama-3.2-11B-V-I
& 54.60
& 20.74
& 4.81
& 59.72
& 33.52
& \scoredelta{173.39}{10.38}
& 49.38
& 18.01
& 1.94
& 47.41
& 16.11
& \scoredelta{132.85}{4.80}
& \scoredelta{306.24}{5.70} \\

InternVL3-8B
& \underline{64.21}
& 38.43
& \underline{16.76}
& \textbf{69.58}
& \textbf{48.43}
& \scoredelta{\underline{237.41}}{1.58}
& 50.86
& 22.69
& 1.76
& 50.93
& 15.74
& \scoredelta{141.98}{0.80}
& \scoredelta{379.39}{2.38} \\

Qwen2.5-VL-7B
& 64.10
& \textbf{41.30}
& 16.29
& 66.57
& 43.52
& \scoredelta{231.78}{6.48}
& \textbf{67.45}
& \textbf{47.45}
& \textbf{21.95}
& \textbf{70.70}
& \textbf{50.56}
& \scoredelta{\textbf{258.11}}{4.05}
& \scoredelta{\textbf{489.89}}{9.85} \\

LLaVA-OneVision-7B
& 58.36
& 32.36
& 10.93
& 62.78
& 36.29
& \scoredelta{200.72}{5.99}
& 52.36
& 27.17
& 4.26
& 59.26
& 29.72
& \scoredelta{172.77}{6.18}
& \scoredelta{373.49}{6.16} \\

Mantis-8B-Idefics2
& 55.21
& 28.29
& 7.96
& 56.94
& 28.98
& \scoredelta{177.38}{4.11}
& 58.94
& 35.05
& 11.76
& 58.93
& 33.70
& \scoredelta{198.38}{5.24}
& \scoredelta{375.76}{7.04} \\

Deepseek-VL-7B
& 58.17
& 29.30
& 7.22
& 60.69
& 27.96
& \scoredelta{183.35}{3.36}
& 50.28
& 20.14
& 0.93
& 49.31
& 8.24
& \scoredelta{128.89}{3.58}
& \scoredelta{312.24}{2.29} \\

Qwen2-VL-7B
& 61.07
& 35.18
& 12.87
& 61.99
& 37.87
& \scoredelta{208.98}{8.73}
& \underline{61.97}
& \underline{37.92}
& 14.26
& 58.71
& 31.30
& \scoredelta{204.16}{2.05}
& \scoredelta{\underline{413.14}}{10.21} \\

LLaVA-NeXT-I-7B
& 56.85
& 30.05
& 8.34
& 60.14
& 33.05
& \scoredelta{188.43}{7.60}
& 61.02
& 37.31
& \underline{15.09}
& \underline{62.69}
& \underline{38.33}
& \scoredelta{\underline{214.44}}{9.44}
& \scoredelta{402.87}{15.00} \\

\bottomrule

\end{tabular}
}

\caption{
Performance comparison of recent closed and open-source LVLMs on MVH-Bench.
The best and second-best results are highlighted in \textbf{bold} and \underline{underline}, respectively.
All results are averaged over three independent runs, with standard deviations reported after $\pm$.
}
\label{tab:baseline-a}

\end{table*}

%% file: tables/baseline_bias.tex
\begin{table}[t]
\centering
\scriptsize
\setlength{\tabcolsep}{5pt} 
\renewcommand{\arraystretch}{1.0}
\resizebox{\columnwidth}{!}{
\begin{tabular}{l|cc|cc}
\toprule
\multirow{3}{*}{\textbf{LVLMs}} 
  & \multicolumn{2}{c|}{\textbf{Cross-Instance}} 
  & \multicolumn{2}{c}{\textbf{Cross-View}} \\
\cmidrule(lr){2-3} \cmidrule(lr){4-5}
  & \textbf{YER} & \textbf{AER}
  & \textbf{YER} & \textbf{AER} \\
\midrule
\multicolumn{5}{c}{\textbf{Closed-Source}} \\
\midrule
GPT-4o & 70.09 & 22.10 & 73.51 & 25.42 \\
Gemini 2.5 Flash & 52.14 & 66.67 & 41.47 & 73.71 \\
Claude Sonnet 4.5 & 39.05 & 64.58 & 35.04 & 76.50 \\
\midrule
\multicolumn{5}{c}{\textbf{Open-Source}} \\
\midrule
InternVL3-14B & 74.64 & 73.26 & 68.72 & 82.60 \\
Gemma3-12B & 86.54 & 75.25 & 76.11 & 72.51 \\
Qwen2.5-VL-7B & 45.50 & 86.98 & 54.80 & 94.13 \\
LLaVA-OneVision-7B & 74.02 & 97.88 & 70.18 & 99.24 \\
LLaVA-NeXT-I-7B & 71.54 & 97.50 & 68.62 & 98.55 \\
\bottomrule
\end{tabular}
}
\caption{Comparison of bias evaluation metrics for recent closed and open-source LVLMs on MVH-Bench.}
\label{tab:yer_aer}
\end{table}

%% file: tables/method.tex
\begin{table*}[tb]

\centering
\scriptsize

\setlength{\tabcolsep}{1.0pt}
\renewcommand{\arraystretch}{1.25}

\resizebox{\textwidth}{!}{%
\begin{tabular}{@{}
>{\raggedright\arraybackslash}p{1.8cm}
!{\vrule width 0.9pt}
>{\centering\arraybackslash}p{0.82cm}
>{\centering\arraybackslash}p{0.82cm}
>{\centering\arraybackslash}p{0.82cm}
|>{\centering\arraybackslash}p{0.82cm}
>{\centering\arraybackslash}p{0.82cm}
!{\vrule width 0.5pt}
>{\centering\arraybackslash}p{1.3cm}
!{\vrule width 0.9pt}
>{\centering\arraybackslash}p{0.82cm}
>{\centering\arraybackslash}p{0.82cm}
>{\centering\arraybackslash}p{0.82cm}
|>{\centering\arraybackslash}p{0.82cm}
>{\centering\arraybackslash}p{0.82cm}
!{\vrule width 0.5pt}
>{\centering\arraybackslash}p{1.3cm}
!{\vrule width 0.9pt}
>{\centering\arraybackslash}p{1.30cm}
@{}}

\toprule

\multirow{3}{*}{\textbf{LVLMs}}
&
\multicolumn{6}{c!{\vrule width 0.9pt}}
{\textbf{Cross-Instance}}
&
\multicolumn{6}{c!{\vrule width 0.9pt}}
{\textbf{Cross-View}}
&
\multirow{3}{*}{\makecell{\textbf{MVH}\\\textbf{Score}}}
\\

\cmidrule(lr){2-7}
\cmidrule(lr){8-13}

&
\multicolumn{3}{c|}{\textbf{Binary}}
&
\multicolumn{2}{c!{\vrule width 0.5pt}}{\textbf{M.C.}}
&
\multirow{2}{*}{\textbf{Score}}
&
\multicolumn{3}{c|}{\textbf{Binary}}
&
\multicolumn{2}{c!{\vrule width 0.5pt}}{\textbf{M.C.}}
&
\multirow{2}{*}{\textbf{Score}}
&
\\

&
\textbf{Acc}
&
\textbf{p-Acc}
&
\textbf{q-Acc}
&
\textbf{Acc}
&
\textbf{p-Acc}
&
&
\textbf{Acc}
&
\textbf{p-Acc}
&
\textbf{q-Acc}
&
\textbf{Acc}
&
\textbf{p-Acc}
&
&
\\

\midrule

\multicolumn{14}{c}{\textbf{LLaVA-OneVision-7B}} \\

\midrule

Base model
& 58.36
& 32.36
& 10.93
& 62.78
& 36.29
& \scoredelta{200.72}{5.99}
& 52.36
& 27.17
& \underline{4.26}
& 59.26
& \textbf{29.72}
& \scoredelta{172.77}{6.18}
& \scoredelta{373.49}{6.16}
\\

VCD
& \underline{60.60}
& 32.59
& \underline{11.94}
& \textbf{68.29}
& 43.98
& \scoredelta{217.40}{3.04}
& \underline{55.07}
& \underline{29.31}
& 2.59
& \underline{61.62}
& 28.43
& \scoredelta{\underline{177.02}}{1.86}
& \scoredelta{\underline{394.42}}{1.34}
\\

ICD
& 60.58
& \underline{33.19}
& 11.30
& 68.06
& \textbf{44.81}
& \scoredelta{\underline{217.94}}{2.49}
& 54.89
& 28.47
& 3.61
& 60.42
& 27.87
& \scoredelta{175.26}{4.30}
& \scoredelta{393.20}{5.95}
\\

AvisC
& 60.46
& 29.31
& 10.28
& \underline{68.15}
& \underline{44.35}
& \scoredelta{212.54}{3.49}
& 53.47
& 26.43
& 2.04
& \textbf{61.81}
& \underline{28.61}
& \scoredelta{172.36}{1.86}
& \scoredelta{384.90}{2.39}
\\

\textbf{RSCD~(Ours)}
& \textbf{62.34}
& \textbf{38.75}
& \textbf{14.35}
& 67.31
& 43.61
& \scoredelta{\textbf{226.36}}{4.43}
& \textbf{56.74}
& \textbf{33.06}
& \textbf{5.65}
& 60.79
& 27.96
& \scoredelta{\textbf{184.19}}{1.99}
& \scoredelta{\textbf{410.55}}{6.41}
\\
\midrule

\multicolumn{14}{c}{\textbf{Qwen2.5-VL-7B}} \\

\midrule

Base model
& \underline{64.10}
& \underline{41.30}
& \underline{16.29}
& {66.57}
& {43.52}
& \scoredelta{\underline{231.78}}{6.48}
& \underline{67.45}
& \underline{47.45}
& \underline{21.95}
& \underline{70.70}
& \underline{50.56}
& \scoredelta{\underline{258.11}}{4.05}
& \scoredelta{\underline{489.89}}{9.85}
\\

VCD
& 62.75
& 38.52
& 13.52
& 67.87
& 44.72
& \scoredelta{227.38}{23.72}
& 66.04
& 45.51
& 19.63
& 70.60
& 46.48
& \scoredelta{248.27}{24.03}
& \scoredelta{475.64}{47.62}
\\

ICD
& 61.69
& 32.13
& 10.46
& 65.55
& 42.04
& \scoredelta{211.88}{19.12}
& 65.77
& 39.31
& 16.39
& 69.67
& 47.41
& \scoredelta{238.54}{26.04}
& \scoredelta{450.42}{44.95}
\\

AvisC
& 61.81
& 36.43
& 11.39
& \underline{68.84}
& \underline{45.93}
& \scoredelta{224.39}{15.60}
& 64.44
& 38.66
& 16.02
& 69.35
& 45.65
& \scoredelta{234.12}{17.49}
& \scoredelta{458.51}{33.08}
\\

\textbf{RSCD~(Ours)}
& \textbf{67.50}
& \textbf{45.19}
& \textbf{17.69}
& \textbf{73.66}
& \textbf{53.70}
& \scoredelta{\textbf{257.73}}{2.90}
& \textbf{72.87}
& \textbf{56.02}
& \textbf{31.02}
& \textbf{77.36}
& \textbf{58.33}
& \scoredelta{\textbf{295.60}}{0.34}
& \scoredelta{\textbf{553.33}}{2.73}
\\

\bottomrule

\end{tabular}%
}

\caption{
Performance of various methods on MVH-Bench using LLaVA-OneVision-7B and Qwen2.5-VL-7B.
The best and second-best results are highlighted in \textbf{bold} and \underline{underline}, respectively.
All results are averaged over three independent runs, with standard deviations reported after $\pm$.
}

\label{tab:method}

\end{table*}

%% file: figures/3_1_analysis.tex
\begin{figure}[t!]
  \centering
    \includegraphics[width=0.98\linewidth]{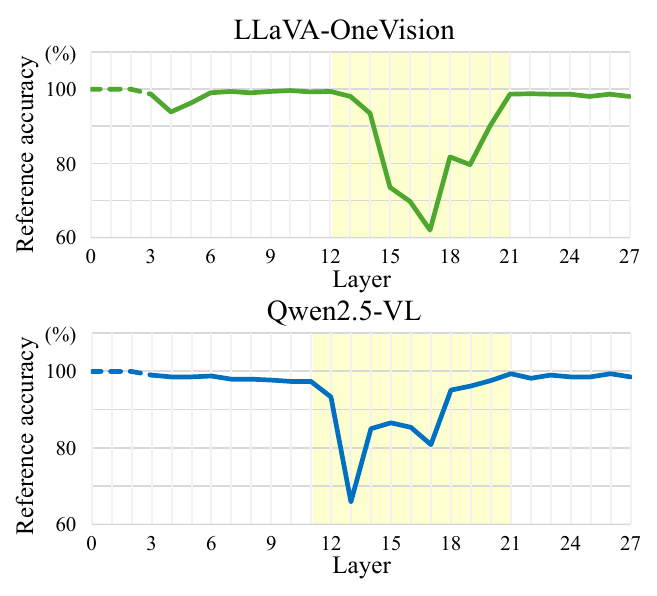}
  \caption{
  Layer-wise analysis of text-to-text attention blocking for LLaVA-OneVision-7B and Qwen2.5-VL on a captioning task.
    The yellow shaded region indicates the layer range selected by RSCD.
  }
  \label{fig:interaction-window}
  
\end{figure}

%% file: sections/5_experiment.tex
\section{Experimental Results}

\subsection{LVLMs' Performance on MVH-Bench}
\label{sec:baseline_experiments}

To assess multi-view hallucination in recent LVLMs, we evaluate five closed-source and ten open-source LVLMs on MVH-Bench (see Table~\ref{tab:baseline-a}).
All models achieve consistently low p-Acc and q-Acc across all categories, indicating that they struggle to resolve the MVH problem.
Furthermore, the models that performed best differed between the Cross-Instance and Cross-View categories, suggesting that these two settings pose distinct challenges.

To quantify the extent to which models are affected by distracting visual cues in MVH-Bench, we report YER and AER in Table~\ref{tab:yer_aer} (see Appendix~\ref{appendix:bias} for results on the remaining models).
Most models exhibit high YER and AER scores, indicating a strong tendency to answer ``Yes'' to binary questions and to select adversarial options in multiple-choice questions.
Notably, open-source models show even higher AER than their closed-source counterparts.
These results suggest that current LVLMs are highly susceptible to multi-view interference.

\subsection{Overview of Baseline Methods}
To evaluate the effectiveness of RSCD, we compare it with recent representative contrastive decoding-based hallucination mitigation approaches, namely VCD~\cite{leng2024vcd}, ICD~\cite{wang2024ICD}, and AvisC~\cite{woo2024dont}.
While all three follow the same idea of constructing a hallucination-prone \emph{negative logits} distribution, they differ in how they obtain the negative logits.
VCD deliberately corrupts the visual input with noise, weakening the model’s access to accurate visual signals and thereby inducing a visually misaligned distribution.
ICD appends role prefixes to the instruction to form disturbance instructions, thereby amplifying hallucinations.
AvisC constructs its negative distribution by conditioning only on image tokens that receive high attention yet are semantically uninformative or query-irrelevant.

\subsection{Performance Evaluation of RSCD}
We assess the effectiveness of RSCD by comparing it with recent hallucination mitigation methods, namely VCD, ICD, and AvisC, on LLaVA-OneVision-7B and Qwen2.5-VL-7B (see Table~\ref{tab:method}).
RSCD consistently outperforms all competing methods on both models.
In particular, it outperforms AvisC by 25.7 points on LLaVA-OneVision-7B and by 94.8 points on Qwen2.5-VL-7B.
Moreover, improvements are observed across most metrics and categories.
These results suggest that RSCD effectively mitigates MVH by encouraging LVLMs to focus on the queried instance or view.

\section{Analysis}

\subsection{Probing the Causes of MVH}
To further investigate whether MVH arises from visual interference, we modify the MVH-Bench validation set by replacing the distractors. 
Specifically, we replace the descriptor phrases in questions whose correct answer is “No” with alternatives that do not appear in the images.
As shown in Figure~\ref{fig:ablation}, the MVH Score increases substantially, while both YER and AER decrease. 
This demonstrates that the poor performance of current models on MVH-Bench is largely driven by visual distraction.

\input{appendix/figures/MVH_cause}

\subsection{Analysis of Layer Range Selection}
To examine the effectiveness of the selected layer range in RSCD, we analyze the results obtained by applying the mask to different layer ranges (see Figure~\ref{fig:hyperparameter}(a)). 
Specifically, using $\Lambda^{\star}$ (the layer range 13--20 in LLaVA-OneVision) as a reference, we evaluate the performance of ranges that precede and follow $\Lambda^{\star}$, as well as the full set of layers.
RSCD achieves the best performance when masking is applied at $\Lambda^{\star}$, and still shows slight improvement over the base model even when extended to later layers.
In contrast, applying RSCD to early layers significantly degrades performance. 

In addition, we analyze the sensitivity of RSCD to the selected layer range by varying the interval around $\Lambda^{\star}$ (see Figure~\ref{fig:hyperparameter}(b)). 
Starting from $\Lambda^{\star}$, we reduce or extend the range by one layer and measure the resulting MVH Score. 
The performance remains stable across these variations, indicating that RSCD is not highly sensitive to the exact layer boundaries.

\subsection{Ablation Study on Hyperparameters}
\input{figures/4_ablation}
We analyze the effect of the two RSCD hyperparameters, $\alpha$ and $\rho$.
As shown in Figure~\ref{fig:hyperparameter}(c), RSCD consistently outperforms the baseline across a wide range of hyperparameter values.
This indicates that RSCD is insensitive to the choice of hyperparameters.
We also observe a clear trend: as the masking ratio $\rho$ increases, performance gradually improves.
This indicates that stronger perturbations to the text tokens produce negative logits that are further shifted away from the correct prediction.
However, in the extreme case where all tokens are masked ($\rho = 1.0$), performance drops substantially. 
This shows that partially masking attention over the text tokens is more effective than complete masking for inducing partially grounded responses.

\subsection{Inference Cost of RSCD}

Conventional contrastive decoding methods for mitigating visual hallucination generate negative logits by modifying the image itself, altering the preceding prompt, or masking a subset of visual tokens.
Such approaches incur additional computational overhead due to input manipulation and the recomputation of intermediate visual representations across transformer layers.
In contrast, RSCD produces negative logits without altering the image input or its token embeddings, thereby enabling direct reuse of cached image key-value states.
Due to this feature, RSCD is particularly efficient in multi-view settings where the input consists of a substantial number of image tokens (see Table~\ref{tab:inference_cost}).
\input{appendix/tables/inference_cost}

\subsection{Qualitative Analysis of RSCD}
\label{sec:qualitative}
To investigate the effect of the reference shift mask, we visualize the changes in the text-to-image attention maps before and after applying the mask (see Figure~\ref{fig:qualitative}).
Specifically, we focus on the attention from the token corresponding to the phrase \textit{``to the right of the person''}. 
To answer the question correctly, the model must verify whether \textit{``the sink''} is actually to the right of the person.
In the top row (base model), the token attends strongly to the sink region (solid yellow line).
In contrast, in the bottom row (with reference shift mask), attention to the sink is suppressed and instead shifts toward the image region corresponding to the phrase \textit{``to the right of the person''} (solid white line).
This shift shows that our masking strategy perturbs the full query context, causing the model to rely on a partially interpreted phrase and resulting in an incorrect answer.
For additional qualitative examples, please refer to Appendix~\ref{appendix:qualitative}.

\input{figures/5_qualitative}

\section{Conclusion}
In this work, we identified multi-view hallucination as a key challenge for LVLMs and introduced MVH-Bench to systematically evaluate their robustness against cross-instance and cross-view interference. 
Experiments on recent LVLMs show that MVH remains a persistent challenge.
To address this issue, we proposed RSCD, a simple yet effective training-free decoding method that suppresses predictions driven by incomplete query understanding.
Extensive experiments demonstrate that RSCD consistently mitigates MVH and outperforms existing hallucination mitigation methods.

%% file: appendix/figures/MVH_cause.tex
\begin{figure}[t!]
  \centering
    \includegraphics[width=\linewidth]{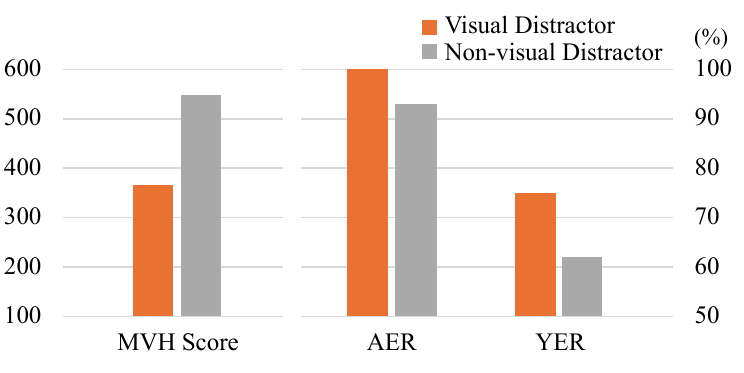}
  \caption{
Effect of replacing visual distractors with random/non-visual distractors on MVH-Bench performance.}
  \label{fig:ablation}
    \vspace{-10pt}
\end{figure}

%% file: figures/4_ablation.tex
\begin{figure}[t!]
  \centering
    \includegraphics[width=\linewidth]{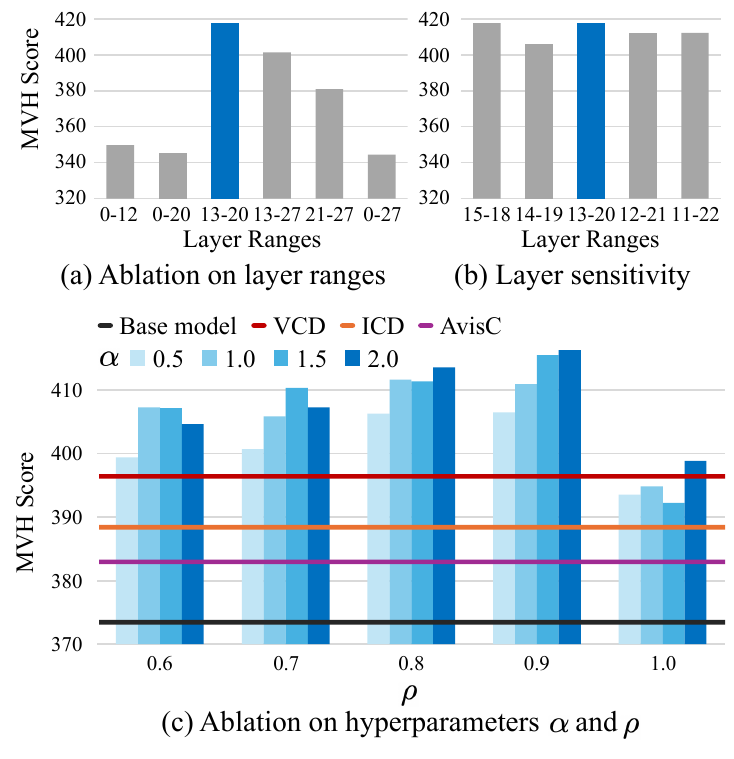}
\vspace{-0.8cm}
  \caption{Analysis of RSCD across different layer ranges and hyperparameter settings.}
\vspace{-0.2cm}
  \label{fig:hyperparameter}
\end{figure}

%% file: appendix/tables/inference_cost.tex
\begin{table}[t]
\centering
\small
\setlength{\tabcolsep}{4pt}
\begin{tabular}{l c}
\toprule
\textbf{Method} & \textbf{Time (sec. /query)} \\
\midrule
Base model & 0.81 \\
VCD & 1.38 \\
ICD & 1.39 \\
AvisC & 1.37 \\
\midrule
RSCD (Ours) & 0.89 \\
\bottomrule
\end{tabular}
  \vspace{-6pt}
\caption{Inference cost comparison between RSCD and other methods used in our experiments.}
\label{tab:inference_cost}
  \vspace{-16pt}
\end{table}

%% file: figures/5_qualitative.tex
\begin{figure}[t!]
\centering
\includegraphics[width=0.91\linewidth]{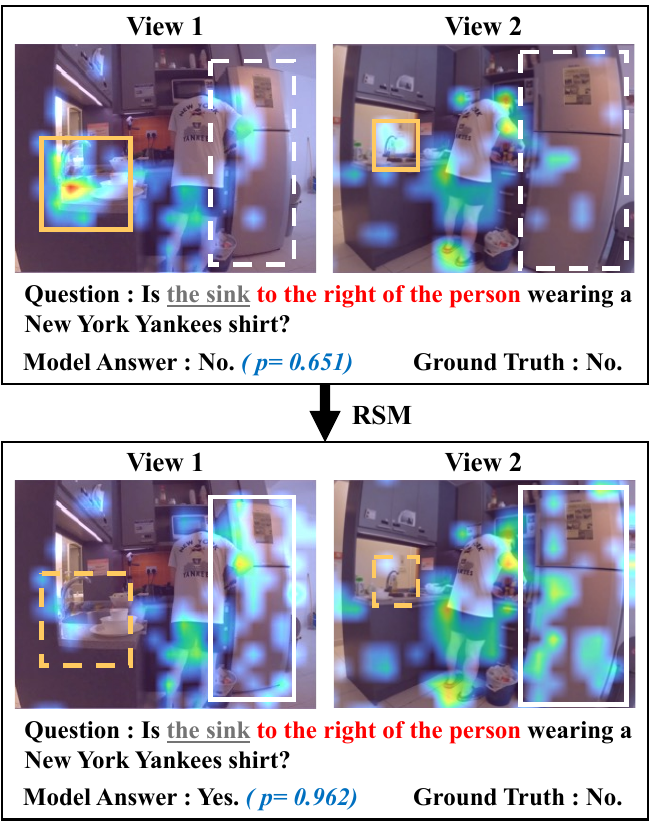}
  \vspace{-5pt}

\caption{Qualitative comparison of attention maps with and without reference shift mask.
Top: attention map for the original prediction.
Bottom: attention map after applying the reference shift mask.
}
\label{fig:qualitative}
  \vspace{-13pt}
\end{figure}

%% file: sections/X_suppl.tex
\clearpage

\input{sections/2_related_work}
\input{appendix/tables/model_comp}

\section{Experimental Details}
\subsection{LVLMs and Experimental Setup}
We evaluate both closed-source and open-source LVLMs in our experiments. 
For closed-source models, we include GPT-5.5~\cite{openai2026gpt55}, GPT-5.4~\cite{openai2026gpt54}, GPT-5.4-mini~\cite{openai2026gpt54}, GPT-4o~\cite{hurst2024gpt}, GPT-4o-mini~\cite{hurst2024gpt}, Gemini-3-Flash~\cite{googledeepmind2025gemini3flash}, Gemini 2.5 Flash~\cite{comanici2025gemini}, Gemini 2.0 Flash~\cite{gemini2.0}, Claude Sonnet 4.6~\cite{anthropic2026claudesonnet46}, and Claude Sonnet 4.5~\cite{claude4.5}. 
For open-source LVLMs, we evaluate the models summarized in Table~\ref{tab:vlm_comparison},  which provides an overview of their architectures, including vision encoders and LLM backbones.
By evaluating a diverse set of vision-language architectures, we can thoroughly assess the robustness of recent LVLMs and their ability to handle multi-view hallucinations.
Experiments conducted with three independent runs are reported with the mean and standard deviation, while those conducted with a single run are reported as a single result.
All evaluations use each model’s default generation settings.
For the RSCD experiments on LLaVA-OneVision-7B and Qwen2.5-VL-7B, we set $(\alpha, \rho, \Lambda^{\star})$ to $(1.0, 0.7, \{13, \dots, 20\})$ and $(1.5, 0.8, \{12, \dots, 20\})$ respectively.
For Qwen3-VL-8B, InternVL3.5-14B, and Llama-3.2-11B-Vision-Instruct, we set $(\alpha, \rho, \Lambda^{\star})$ to $(1.0, 0.8, \{11, \dots, 25\})$, $(1.0, 0.8, \{13, \dots, 29\})$, and $(1.5, 0.9, \{5, \dots, 23\})$, respectively.
Unless otherwise specified, all analyses are conducted with LLaVA-OneVision-7B and run on NVIDIA RTX A6000 GPUs.

\subsection{Details of the Captioning Analysis} \label{appendix:captioning}
In this section, we provide additional details on the captioning task described in Section~\ref{3.2}.
Here, we adopt a captioning-based analysis to explicitly examine which image content the model relies on when generating its response.
Specifically, we construct image pairs using samples from the COCO dataset~\cite{lin2014microsoft} and generate captions using the prompt \textit{``Provide a one-sentence caption for Image 1/2''}. 
We primarily use COCO images for this analysis because the high degree of visual similarity across views in MVH-Bench makes it difficult to clearly attribute the generated response to a specific visual input.
When blocking text-to-text attention in intermediate layers, we observe that the model often generates captions describing the non-target image or produces sentences that mix information from both images. 
These behaviors indicate that disrupting the formation of textual context can cause the model to rely on irrelevant visual information.

In the main paper, we introduce a metric called \textit{reference accuracy} that assesses whether the generated caption is well grounded in the target image. 
To compute this metric, we first obtain reference captions by prompting the model with only a single image. 
We then compare the generated caption with the reference captions using CLIP embedding similarity and consider the prediction correct if it is more similar to the reference caption of the target image.

\input{appendix/tables/recent_model}
\input{appendix/tables/distinct_failure}
\input{appendix/tables/baseline_bias_appendix}

\input{appendix/tables/paraphrase}

\section{Additional Experiments on MVH-Bench}

\subsection{Evaluation of Recently Released LVLMs on MVH-Bench}
We additionally evaluate recently released LVLMs on MVH-Bench, with the results presented in Table~\ref{tab:recent_baseline}.
While newer models generally achieve higher MVH Scores than the models originally evaluated in the main paper, they still struggle with MVH.
These results indicate that MVH remains a challenge for recent LVLMs and also underscore the importance of MVH-Bench as a diagnostic benchmark.

\subsection{Comparison with Conventional Hallucination Benchmarks}
\label{appendix:distinct_failure_mode}
To examine the relationship between conventional hallucination and MVH, we compare model performance across standard hallucination benchmarks (POPE~\cite{li2023POPE}, CHAIR~\cite{rohrbach2019chair}) and MVH-Bench.
As shown in Table~\ref{tab:hallucination_comparison}, we observe that the model rankings are not consistent across these benchmarks. 
For example, InternVL3-8B achieves the best performance on conventional hallucination metrics (e.g., highest POPE accuracy and lowest CHAIR scores), but does not achieve the best performance on MVH-Bench. 
Conversely, Qwen2.5-VL-7B achieves the highest MVH Score, despite showing relatively lower performance among the compared models on both POPE and CHAIR.
These discrepancies indicate that lower hallucination rates on conventional benchmarks do not directly translate to better MVH performance, suggesting that MVH captures a distinct failure mode requiring dedicated evaluation.

\subsection{Additional Results on Bias Evaluation} \label{appendix:bias}
We additionally report YER and AER for the remaining models not covered in the main text (see Table~\ref{tab:bias_appendix}).

\subsection{Effect of Query Paraphrasing}
To examine the effect of query paraphrasing on MVH-Bench, we paraphrase the validation-set questions using GPT-4o and evaluate model performance.
As shown in Table~\ref{tab:paraphrase}, paraphrasing affects model performance, but the direction and magnitude of the change vary across models.
Nevertheless, MVH is consistently observed under both the original and paraphrased queries, suggesting that the phenomenon is not specific to a particular wording of the benchmark questions.

\input{appendix/figures/ko_analysis_appendix}

\section{Further Analysis of Text-to-Text Attention Masking}
\label{appendix:t2t_analysis}

\subsection{Text-to-Text Attention Masking in a Text-Only Setting}
\label{appendix:text_only}

Our text-to-text attention masking analysis on captioning tasks suggests that text-to-text attention plays an important role in visual grounding. 
However, this observation alone may not be sufficient to conclude that the effect is caused by incomplete textual understanding, since textual interpretation and visual grounding are tightly coupled in multimodal question answering.

To further substantiate our findings, we conduct an analysis in a text-only setting to isolate textual reasoning from multimodal interactions. 
Specifically, we perform a layer-wise text-to-text attention blocking experiment with LLaVA-OV-7B on HotpotQA~\cite{yang2018hotpotqa}, a benchmark that requires multi-hop reasoning over evidence distributed across multiple paragraphs. 
We evaluate the quality of the predicted answer and supporting facts generated by the model, using exact match (EM) for answer accuracy and the F1 score (Sup-F1) for supporting facts. 
Notably, as shown in Table~\ref{tab:text_only}, Sup-F1 drops more sharply than EM and remains low in the middle layer range (13--20). 
This pattern is consistent with the captioning analysis presented in the main paper. 
The discrepancy between EM and Sup-F1 suggests that, even when producing the same answer, the model may rely on incorrect yet superficially relevant evidence. 
This reflects a reduced ability to distinguish gold supporting facts from partially related cues. 
Taken together, these results provide empirical support that masking text-to-text attention disrupts complete textual understanding and causes the model to rely more on partial matches.

\input{appendix/tables/knockout_text_only}
\input{appendix/tables/mvh_captioning_ko}

\input{appendix/tables/rscd_generalization}

\subsection{Text-to-Text Attention Masking across Architectures and Model Scales}

To demonstrate the generality of our findings, we extend the layer-wise text-to-text masking analysis to three additional models\footnote{Llama-3.2-11B-V-I integrates visual and textual modalities through dedicated cross-attention layers, whereas Qwen3-VL and InternVL3.5 concatenate visual tokens with text tokens and perform joint processing via self-attention.}: Qwen3-VL-8B-Instruct, InternVL3.5-14B, and Llama-3.2-11B-Vision-Instruct (see Figure~\ref{fig:ko_analysis_appendix}).
Despite differences in model scale, training pipeline, and multimodal integration strategy, all three models exhibit a consistent U-shaped pattern in reference accuracy across the middle layers.
This consistency, together with the results in the main paper, suggests that the observed behavior is not a heuristic assumption, but an empirically grounded phenomenon.

\subsection{Text-to-Text Attention Masking on MVH-Bench}

We additionally perform the same layer-wise text-to-text attention blocking analysis on the MVH-Bench validation subset.
In MVH-Bench, the two views often contain highly overlapping visual information, such that reference captions generated from each view can include similar objects or descriptions.
Moreover, even without blocking text-to-text attention, the model sometimes generates hallucinated descriptions of the non-queried view, making it difficult to isolate the effect of text-to-text masking.
Despite these limitations, we observe the same trend: while the overall reference accuracy is lower, it drops clearly in the middle layers, particularly around center layers 13--20 (see Table~\ref{tab:mvh_captioning_ko}).

\section{Additional Evaluation and Analysis of RSCD}

\subsection{Generalization Across Model Families and Scales}
\label{appendix:recently_released_lvlms}
To further validate the generalizability of RSCD, we evaluate it across recent LVLMs with diverse model families, architectures, and scales (Qwen3-VL-8B, InternVL3.5-14B, and Llama-3.2-11B-Vision-Instruct). 
As shown in Table~\ref{tab:rscd_generalization}, we observe that RSCD consistently outperforms competing methods across all evaluated models, demonstrating its effectiveness and generalizability.

\subsection{Generalization Across Other Multimodal Benchmarks}
To assess the generalization capability of RSCD, we additionally evaluate it on widely used multimodal benchmarks, including MMBench~\cite{mmbench}, SEED-Bench\footnote{Due to its large scale, we randomly sample 50 examples from each of the 12 categories, including 3 video categories.}~\cite{li2023seedbench}, POPE, and CHAIR, using LLaVA-OV-7B. 
Note that we use the same hyperparameters for RSCD as reported in the manuscript across all benchmarks. 
We do not perform any task-specific tuning.

As shown in \Cref{tab:generalization_mm,tab:generalization_pope,tab:generalization_chair}, RSCD improves performance on most benchmarks, demonstrating its effectiveness across diverse multimodal evaluation settings.
The only exception is CHAIRs, where RSCD shows a slight performance degradation.
RSCD generates captions with more object mentions per caption (from 8.78 to 10.58 object mentions per caption).
Since CHAIRs is a sentence-level metric that marks a caption as incorrect if it contains any hallucinated object, mentioning more objects increases the chance that a caption is counted as hallucinated.
Notably, the improved CHAIRi and Recall indicate that RSCD reduces the proportion of hallucinated objects while capturing more visual content.

\input{appendix/tables/generalization_other_benchmark}

\subsection{Evaluation Beyond Two-View Settings} 
To examine MVH in settings with more than two views, we conduct an additional evaluation on the MVH-Bench validation set by using all available synchronized multi-view images from the source dataset. 
We denote this setting as the max-view setting, where the average number of input images is 4.3.

As shown in Table~\ref{tab:max_view}, compared to the original two-view setting, we observe performance degradation under the max-view setting in both cross-instance and cross-view categories, indicating that MVH becomes more challenging as the number of views increases. 
Using the same hyperparameters as those used in the main paper, RSCD achieves the best overall performance among decoding-based methods, demonstrating its robustness as the number of input views increases.

\input{appendix/tables/max_view}

\subsection{Evaluation of Free-Form Generation}
\label{appendix:free_form}

To examine MVH beyond constrained question-answering formats, we conduct an additional captioning evaluation with LLaVA-OV-7B on the MVH-Bench validation set.
Given multi-view images, we ask the model to generate a caption for a queried view.
As shown in Table~\ref{tab:free_form_captioning}, the base model often grounds its captions in non-queried views, resulting in relatively low reference accuracy.
When RSCD is applied, reference accuracy improves from 69\% to 77\% without reducing the response length.
In addition, when we reformulate MVH-Bench into an open-ended format, RSCD outperforms the base model (Table~\ref{tab:open_ended}).
This suggests that the MVH problem goes beyond rigid QA formats and that RSCD remains effective in free-form generation.

\input{appendix/tables/free_form_captioning}
\input{appendix/tables/open_ended}

\subsection{Analysis of Layer Range Selection Cost}\label{appendix:calibration_cost}
RSCD requires a lightweight layer selection procedure to identify the layer range $\Lambda^{\star}$ for each model.
In our experiments, we perform this selection by probing only 30 paired-image samples, generating a short caption for each image.
Notably, the middle-layer degradation pattern is clearly observable with as few as 3 paired-image samples, suggesting that the layer selection process can be performed with much lower cost if needed.
Furthermore, experiments across diverse datasets and layer-wise analyses show that RSCD remains effective within the selected layer range for each model and is robust to minor variations in the selection.
Therefore, once the layer range is identified, RSCD does not require dataset-specific calibration.

\subsection{Additional Qualitative Examples} \label{appendix:qualitative}
In Figure~\ref{fig:qualitative_appendix}, we present additional qualitative examples which are consistent with the observations in Section~\ref{sec:qualitative}.
As shown in the left part of Figure~\ref{fig:qualitative_appendix}, the baseline model attends primarily to the \textit{``object located in the upper left of the image''} in \textit{``view 1''} (solid yellow line).
However, after applying RSM, the attention spreads to the corresponding object in both views.
Similarly, in the right part of Figure~\ref{fig:qualitative_appendix}, the baseline model correctly attends to the \textit{``person in a gray shirt''} when answering whether they are \textit{``standing with their hands behind their back''} (solid yellow line).
When RSM is employed, the model shifts its attention to another person who is \textit{``standing with their hands behind their back''}, rather than the referenced \textit{``person in a gray shirt''}.
These examples clearly illustrate that RSM distorts the intended semantics of the query, so that the model might attend to semantically mismatched regions.

\section{Additional Details of MVH-Bench}
\subsection{Benchmark Statistics}
Figure~\ref{fig:statistics} summarizes the statistics of MVH-Bench.
Figure~\ref{fig:statistics}(a) and (b) show the distribution of video categories for image pairs extracted from the Ego-Exo4D and LEMMA datasets, respectively.
Figure~\ref{fig:statistics}(c) illustrates the distribution of image pair view configurations formed by combining first-person view (FPV) and third-person view (TPV): FPV\&TPV, TPV\&TPV, and FPV\&FPV.
Figure~\ref{fig:statistics}(d) presents the source dataset composition, where 66\% of the samples originate from Ego-Exo4D and the remaining 34\% from LEMMA.
These statistics demonstrate that MVH-Bench covers diverse activity scenarios and maintains a balanced distribution of multi-view configurations.

\subsection{Formal Definition of Evaluation Metrics} \label{appendix:eval metrics}
For completeness, we provide the formal definitions of the evaluation metrics introduced in Section~\ref{sec:metric}.  
Let $m(q)$ be the model’s prediction for the question $q$, and $\I[\cdot]$ be the indicator function.
Here, $x,~y \in \{i,~j\}$ denote indices of the sampled instance-descriptor pairs, where $x$ specifies the instance ($I_x$) and $y$ specifies the descriptor ($D_y$).
MVH-Bench contains $N$ multi-view image pairs, with the superscript $n$ indexing the $n$-th image pair ($n \in \{1,\ldots,N\}$).

For binary questions, the mean accuracy Acc, pair accuracy p-Acc, and quadruplet accuracy q-Acc are defined as
\begin{equation}
\text{Acc}
= \frac{1}{4N}\sum_{n,x,y}
\I\!\big[m(q_{xy}^{\,n}) = a_{xy}^{\,n}\big],
\label{eq:bin-avg}
\end{equation}
\begin{equation}
\text{p-Acc}
= \frac{1}{2N}\sum_{n,x}
\prod_{y}
\I\!\big[m(q_{xy}^{\,n}) = a_{xy}^{\,n}\big],
\label{eq:bin-pair}
\end{equation}
\begin{equation}
\text{q-Acc}
= \frac{1}{N}\sum_{n}
\prod_{x,y}
\I\!\big[m(q_{xy}^{\,n}) = a_{xy}^{\,n}\big].
\label{eq:bin-quad}
\end{equation}
We additionally define the yes-error ratio (YER) as
\begin{equation}
\text{YER}
= \frac{
\displaystyle \sum_{n,x,y}
\I\!\big[m(q_{xy}^{\,n})=\text{Yes}\big]\;
\I\!\big[m(q_{xy}^{\,n})\neq a_{xy}^{\,n}\big]
}{
\displaystyle \sum_{n,x,y}
\I\!\big[m(q_{xy}^{\,n})\neq a_{xy}^{\,n}\big]
}.
\label{eq:yer}
\end{equation}

For multiple-choice questions, the mean accuracy Acc and pair accuracy p-Acc are defined as
\begin{equation}
\text{Acc}
= \frac{1}{2N}\sum_{n,(x,y)}
\I\!\big[m(q_{xy}^{\,n}) = a_{xy}^{\,n}\big],
\label{eq:multi-acc}
\end{equation}
\begin{equation}
\text{p-Acc}
= \frac{1}{N}\sum_{n}\prod_{(x,y)}
\I\!\big[m(q_{xy}^{\,n}) = a_{xy}^{\,n}\big].
\label{eq:multi-p-acc}
\end{equation}
Finally, we define the adversarial error ratio (AER) as
\begin{equation}
\text{AER}
= \frac{
\displaystyle \sum_{n,(x,y)}
\I\!\big[m(q_{xy}^{\,n})=\text{B}\big]\;
\I\!\big[m(q_{xy}^{\,n})\neq a_{xy}^{\,n}\big]
}{
\displaystyle \sum_{n,(x,y)}
\I\!\big[m(q_{xy}^{\,n})\neq a_{xy}^{\,n}\big]
}.
\label{eq:adv-error}
\end{equation}

\input{appendix/figures/statistics}

\subsection{Data Construction and Human Verification Details} \label{appendix:human verification}

We provide additional details on the data construction and human verification process introduced in the main paper.

First, in the instance-descriptor pair extraction stage, we obtained 68,298 instance-descriptor pairs from 24,564 images. 
Based on these pairs, the automated QA generation process produced 111,671 QA pairs. 
During human verification, we filtered out questions with ambiguous references, such as those referring to non-specific instances like “background”, “something”, or “item”. 
To improve dataset diversity, we kept at most two QA sets per category for each image pair.

The human verification stage was conducted by five annotators with expertise in vision-language reasoning.
As shown in Figure~\ref{fig:interface}, each annotator was provided with a pair of multi-view images, the instance-descriptor pairs used to generate questions for each view, and the automatically generated QA pairs.
We adopted a hierarchical verification process to ensure annotation quality.
Specifically, four annotators independently verified the QA pairs for one subcategory, while the remaining annotator reviewed the verified QA pairs together with the annotator responsible for the corresponding subcategory.
Through this joint review process, disagreements were resolved, and samples that did not reach consensus were discarded.

During verification, we observed several failure patterns in the QA pairs generated by GPT-4o.
For spatial descriptors, ambiguities frequently arise when the same relation applies to multiple objects; for example, descriptors such as ``on the table'' or ``beside the wall'' may refer to multiple instances.
For action descriptors, overly generic verbs such as ``standing'' or ``looking'' are frequently generated.
In addition, some numerical questions can be answered without visual grounding, such as questions asking for the ``number of hands''.

After verification, the final dataset consists of 4,800 QA pairs and 1,274 images, corresponding to approximately 96\% of the generated samples being filtered out during the human verification stage. 
Among the samples retained in the final benchmark, we estimate that around 70\% involved partial human editing, including refining descriptors, specifying instances more clearly, correcting inaccurate answers, or fixing grammatical issues.

\subsection{Rationale Behind the Template-Based Design of MVH-Bench}\label{appendix:template_rationale}
The goal of MVH-Bench is to provide a controlled diagnostic evaluation of multi-view hallucination. 
By holding the question format constant and varying only the instances, views, and/or descriptors, we ensured that any model errors could be attributed to incorrect instance/view grounding rather than to the question format itself. 
To balance this control with generalizability, we designed diverse templates across categories, subcategories, question types, and perspectives (TPV and FPV).

The purpose of instance-descriptor swapping is to construct visually plausible hard negatives. 
This ensures that success requires precise grounding rather than reliance on superficial cues. 
Each hard negative is paired with its corresponding positive example to enable pairwise evaluation for each question. 
This pairwise evaluation protocol, widely used in visual grounding benchmarks such as POPE and MME, provides a stricter test of consistent grounding and true reasoning.
In MVH-Bench, we further strengthen this evaluation by adding q-Acc, which measures whether the model answers matched quadruples consistently. 
In short, the essence of our approach is to make it harder for models to obtain high scores through shortcuts, such as relying on only one view or part of the multi-view input.

\subsection{Prompts for Instance-Descriptor Pair Extraction} \label{appendix:prompt}
We design tailored prompts for instance-descriptor pair extraction for each subcategory (action, object, numerical, and spatial) and for each viewpoint (FPV and TPV).
\Cref{fig:action_first,fig:object_first,fig:numerical_first,fig:spatial_first} show the prompts used for the first-person viewpoint for the four subcategories, while \Cref{fig:action_third,fig:object_third,fig:numerical_third,fig:spatial_third} illustrate the corresponding prompts for the third-person viewpoint.

\section{Ethics Statement}
To ensure responsible data use, we obtained the necessary licenses from the contributing institutions to use the Ego-Exo4D and LEMMA datasets in this study.

\input{appendix/figures/qualitative}

\input{appendix/figures/human_interface}

\input{appendix/tables/prompt}

%% file: sections/2_related_work.tex
\section{Related Work}
\subsection{Multi-View Datasets and Tasks}
Multi-view images capture complementary visual cues that provide richer information than a single view, leading to a more complete understanding of the scene.
This advantage has motivated the development of diverse multi-view datasets.
Among these, several works have introduced datasets that capture the spatial layout of indoor environments from diverse viewpoints~\cite{silberman2012indoor, chang2017matterport3d, yeshwanth2023scannet++}.

In parallel, other works have introduced datasets that capture real-world human activities and interactions from first and third-person perspectives~\cite{sigurdsson2018charades, jia2020lemma, 10658224egoexo4d}. 
These datasets have been widely used to support various vision tasks, including 3D reconstruction~\cite{yang2025fast3r, cabon2025must3r, wang2025continuous}, depth estimation~\cite{piccinelli2025unidepthv2, guo2025depth}, correspondence matching~\cite{leroy2024grounding, mur2025mama}, and viewpoint transformation~\cite{luo2024put}.

In addition to these applications, these datasets have also been explored in LVLM research, where multi-view information is leveraged to generate responses grounded in extended environmental context and to support human-AI interaction.
One line of research examines whether LVLMs can fully interpret the 3D structure and spatial relationships of environments from multi-view data~\cite{hong20233d, yeh2025seeing}.
Another line of work investigates whether LVLMs can function as visual assistants capable of holistic scene understanding~\cite{EmbodiedQA, islam2023eqa, lee2025towards}.
Despite rapid progress in multi-view understanding, the problem of multi-view hallucination largely remains unaddressed.

\subsection{Hallucination in LVLMs}

Hallucination in LVLMs refers to the generation of responses that are inconsistent with the underlying visual evidence.
To quantitatively assess the susceptibility and robustness of LVLMs to hallucination, several benchmarks have been proposed.
Some benchmarks focus on object-level hallucinations by evaluating whether mentioned objects are actually present in the image~\cite{rohrbach2019chair,li2023POPE}, while others extend this perspective to hallucinations about object attributes and inter-object relationships~\cite{wang2024amber,sun2023MMHAL-Bench,zheng2025reefknot,wu2024Rbench}.
Beyond single-image settings, MIHBench~\cite{li2025mihbench} evaluates a model’s ability to aggregate visual information across multiple images to form a unified judgment. 
More recent works have also considered video settings, introducing datasets that capture dynamic and temporal hallucinations in LVLMs~\cite{choong2024vidhal,li2025vidhalluc,kong2025mhbench,gao2025HAVEN}.
For example, MHBench~\cite{kong2025mhbench} and VidHalluc~\cite{li2025vidhalluc} assess the robustness of reasoning over semantically similar actions or the temporal order of frames. 

Building on these benchmarks, recent mitigation approaches encourage LVLMs to rely more on the visual input when generating responses.
To this end, recent methods aim to directly strengthen LVLMs’ reliance on visual input.
These approaches steer more attention toward image tokens during inference~\cite{yin2025clearsight,he2025cracking,tang2025VisFlow,jiang2025devils}, leverage auxiliary signals to highlight informative regions of the visual input~\cite{li2025vidhalluc}, or refine logits with image-conditioned signals~\cite{an2025AGLA,deng2024clipguidedecoding}.
Other methods mitigate hallucinations by explicitly inducing and then subtracting hallucination-driving factors during decoding.
For example, some methods deliberately amplify language priors or hallucination-inducing visual cues and then remove their contribution at inference time~\cite{leng2024vcd,wang2024ICD,woo2024dont,kong2025mhbench,gong2024damro,kim2024vacode,park2025evaluating}.
Some approaches instead retrospectively verify generated responses and regenerate them when potential hallucinations are detected~\cite{wu2026generatebutverify, dard}.
These benchmarks and methods have advanced the analysis and mitigation of hallucinations in LVLMs, but to the best of our knowledge, hallucinations in multi-view settings remain largely unexplored.

%% file: appendix/tables/model_comp.tex
\setlength{\tabcolsep}{22pt}
\begin{table*}[t!]
    \centering
    \renewcommand{\arraystretch}{1.25}
    \fontsize{17}{17}\selectfont
    \resizebox{0.8\linewidth}{!}{
\begin{tabular}{lcc}
\toprule
\textbf{LVLM} & \textbf{Vision Encoder} & \textbf{LLM Backbone} \\
\midrule
InternVL3-14B~\cite{wang2025internvl3} & InternViT-300M &
Qwen2.5-14B  \\
Gemma3-12B~\cite{team2025gemma} & SigLIP-400M & Gemma3-12B  
\\
Llama-3.2-11B-V-I~\cite{llama32vision} & - & Llama-3.1-8B
  \\
InternVL3-8B~\cite{wang2025internvl3} & InternViT-300M & Qwen2.5-7B \\
Qwen2.5-VL-7B~\cite{bai2025qwen2}  & ViT (customized) & Qwen2.5-7B  \\
LLaVA-OneVision-7B~\cite{li2024llava} & SigLIP-SO & Qwen2-7B \\
Mantis-8B-Idefics2~\cite{jiang2024mantis} & SigLIP & Mistral-7B-v0.1 \\
Deepseek-VL-7B~\cite{lu2024deepseek} & SigLIP-L, SAM-B & DeepSeek-LLM-7B \\
Qwen2-VL-7B~\cite{wang2024qwen2}  & ViT-L & Qwen2-7B \\
LLaVA-NeXT-I-7B~\cite{li2024llavani} & SigLIP-SO & Qwen1.5-7B  \\
InternVL3.5-14B~\cite{wang2025internvl35} & InternViT-300M  &  Qwen3-14B  \\
InternVL3.5-8B~\cite{wang2025internvl35} & InternViT-300M & Qwen3-8B  \\
Qwen3-VL-8B ~\cite{bai2025qwen3vl} & SigLIP2-SO-400M & Qwen3-8B  \\

\bottomrule
\end{tabular}
}
\caption{
Comparison of open-source LVLMs across vision encoders and LLM architectures.}
\label{tab:vlm_comparison}
\end{table*}
\setlength{\tabcolsep}{6pt}

%% file: appendix/tables/recent_model.tex
\begin{table*}[tb]

\centering
\small

\setlength{\tabcolsep}{2pt}
\renewcommand{\arraystretch}{1.25}

\begin{tabularx}{\textwidth}{@{}
>{\raggedright\arraybackslash}p{2.4cm}
!{\vrule width 0.9pt}
*{3}{>{\centering\arraybackslash}X}
!{\vrule width 0.5pt}
*{2}{>{\centering\arraybackslash}X}
!{\vrule width 0.5pt}
>{\centering\arraybackslash}X
!{\vrule width 0.9pt}
*{3}{>{\centering\arraybackslash}X}
!{\vrule width 0.5pt}
*{2}{>{\centering\arraybackslash}X}
!{\vrule width 0.5pt}
>{\centering\arraybackslash}X
!{\vrule width 0.9pt}
>{\centering\arraybackslash}X
@{}}

\toprule

\multirow{3}{*}{\textbf{LVLM}}
&
\multicolumn{6}{c!{\vrule width 0.9pt}}{\textbf{Cross-Instance}}
&
\multicolumn{6}{c!{\vrule width 0.9pt}}{\textbf{Cross-View}}
&
\multirow{3}{*}{\makecell{\textbf{MVH}\\\textbf{Score}}}
\\

\cmidrule(lr){2-7}
\cmidrule(lr){8-13}

&
\multicolumn{3}{c!{\vrule width 0.5pt}}{\textbf{Binary}}
&
\multicolumn{2}{c!{\vrule width 0.5pt}}{\textbf{M.C.}}
&
\multirow{2}{*}{\textbf{Score}}
&
\multicolumn{3}{c!{\vrule width 0.5pt}}{\textbf{Binary}}
&
\multicolumn{2}{c!{\vrule width 0.5pt}}{\textbf{M.C.}}
&
\multirow{2}{*}{\textbf{Score}}
&
\\

&
\textbf{Acc}
&
\textbf{p-Acc}
&
\textbf{q-Acc}
&
\textbf{Acc}
&
\textbf{p-Acc}
&
&
\textbf{Acc}
&
\textbf{p-Acc}
&
\textbf{q-Acc}
&
\textbf{Acc}
&
\textbf{p-Acc}
&
&
\\

\midrule

\multicolumn{14}{c}{\textbf{Closed-Source}} \\

\midrule

GPT-5.5
& 69.10 & 45.83 & 23.06
& 73.89 & 55.56 & \underline{267.43}
& 76.60 & 59.17 & 37.50
& \textbf{82.78} & \textbf{68.33} & \textbf{324.38}
& \textbf{591.81} \\

GPT-5.4
& \underline{70.00} & 44.86 & 24.72
& \textbf{75.28} & \textbf{57.78} & \textbf{272.64}
& 73.33 & 51.81 & 31.11
& \underline{80.69} & \underline{64.44} & \underline{301.39}
& \underline{574.03} \\

GPT-5.4 mini
& 62.36 & 37.92 & 14.44
& 64.72 & 42.50 & 221.94
& 64.03 & 40.69 & 13.61
& 65.28 & 38.33 & 221.94
& 443.89 \\

Gemini 3.0 Flash
& \textbf{75.14} & \underline{57.92} & \underline{35.56}
& 47.50 & 26.67 & 242.78
& \textbf{81.53} & \underline{69.58} & \underline{49.17}
& 58.33 & 39.72 & 298.33
& 541.11 \\

Claude Sonnet 4.6
& 68.96 & \textbf{59.44} & \textbf{38.61}
& 50.56 & 39.17 & 256.74
& \underline{77.08} & \textbf{70.00} & \textbf{53.06}
& 48.89 & 38.61 & 287.64
& 544.38 \\

\midrule

\multicolumn{14}{c}{\textbf{Open-Source}} \\

\midrule

InternVL3.5-14B
& 62.36 & 34.17 & 12.78
& \textbf{75.69} & \underline{58.89} & 243.89
& \underline{70.28} & 47.08 & 25.56
& \textbf{80.83} & \textbf{66.11} & \underline{289.86}
& 533.75 \\

InternVL3.5-8B
& \underline{63.75} & \underline{38.89} & \underline{19.17}
& 73.33 & 55.28 & \underline{250.42}
& \textbf{71.94} & \textbf{51.67} & \underline{27.50}
& \underline{78.75} & \underline{61.39} & \textbf{291.25}
& \underline{541.67} \\

Qwen3-VL-8B
& \textbf{71.46} & \textbf{53.19} & \textbf{28.33}
& \underline{75.69} & \textbf{59.44} & \textbf{288.12}
& 70.76 & \underline{52.92} & \textbf{29.17}
& 76.81 & 58.33 & 287.99
& \textbf{576.11} \\

\bottomrule

\end{tabularx}

\caption{
Performance comparison of recently released LVLMs on MVH-Bench.
The best and second-best results are highlighted in \textbf{bold} and \underline{underline}, respectively.
}

\label{tab:recent_baseline}

\end{table*}

%% file: appendix/tables/distinct_failure.tex
\begin{table*}[t]
\centering
\small
\begin{tabular}{lcccc}
\toprule
\textbf{LVLM} & \textbf{POPE Acc.} $\uparrow$ & \textbf{CHAIRs} $\downarrow$ & \textbf{CHAIRi} $\downarrow$ & \textbf{MVH Score} $\uparrow$ \\
\midrule
Qwen2.5-VL-7B & 86.77 & 47.40 & 13.12 & \textbf{489.89} \\
LLaVA-OV-7B   & 87.38 & 43.20 & 11.07 & 373.49 \\
InternVL3-8B  & \textbf{89.63} & \textbf{31.20} & \textbf{8.80} & 379.39 \\
\bottomrule
\end{tabular}
\caption{Comparison of model performance on conventional hallucination benchmarks and MVH-Bench.}
\vspace{-0.1cm}
\label{tab:hallucination_comparison}
\end{table*}

%% file: appendix/tables/baseline_bias_appendix.tex
\begin{table}[t]
\centering
\scriptsize
\setlength{\tabcolsep}{5pt} 
\renewcommand{\arraystretch}{1.0}
\resizebox{\columnwidth}{!}{
\begin{tabular}{l|cc|cc}
\toprule
\multirow{2}{*}{\textbf{LVLMs}} 
  & \multicolumn{2}{c|}{\textbf{Cross-Instance}} 
  & \multicolumn{2}{c}{\textbf{Cross-View}} \\
\cmidrule(lr){2-3} \cmidrule(lr){4-5}
  & \textbf{YER} & \textbf{AER}
  & \textbf{YER} & \textbf{AER} \\
\midrule
\multicolumn{5}{c}{\textbf{Closed-Source}} \\
\midrule
GPT-4o mini & 82.52 & 68.72 & 67.85 & 71.34 \\
Gemini 2.0 Flash & 72.01 & 61.19 & 73.57 & 64.43 \\
\midrule
\multicolumn{5}{c}{\textbf{Open-Source}} \\
\midrule
Llama-3.2-11B-V-I & 85.94 & 85.95 & 79.75 & 94.51 \\
InternVL3-8B & 76.66 & 81.58 & 71.90 & 88.58 \\
Mantis-8B-Idefics2 & 62.50 & 71.55 & 61.89 & 77.73 \\
Deepseek-VL-7B & 66.46 & 88.47 & 74.75 & 95.86 \\
Qwen2-VL-7B & 66.84 & 91.48 & 61.03 & 95.37 \\
\bottomrule
\end{tabular}
}
\caption{Comparison of bias evaluation metrics for recent closed and open-source LVLMs on MVH-Bench.}
\vspace{-0.2cm}
\label{tab:bias_appendix}
\end{table}

%% file: appendix/tables/paraphrase.tex
\begin{table*}[t]
\centering
\small
\begin{tabular}{llccc}
\toprule
\textbf{Model} & \textbf{Query Type}
& \textbf{Cross-Instance}
& \textbf{Cross-View}
& \textbf{MVH} \\
& & \textbf{Score} & \textbf{Score} & \textbf{Score} \\
\midrule

\multirow{2}{*}{LLaVA-OV-7B}
& Original   & 216.25 & 156.88 & 373.12 \\
& Paraphrase & 211.25 & 173.12 & 384.38 \\
\midrule

\multirow{2}{*}{Qwen2.5-VL-7B}
& Original   & 243.12 & 286.25 & 529.38 \\
& Paraphrase & 225.00 & 248.12 & 473.12 \\

\bottomrule
\end{tabular}
\vspace{-0.1cm}
\caption{Performance of LVLMs on original and paraphrased queries from the MVH-Bench validation set.}
\vspace{-0.1cm}
\label{tab:paraphrase}
\end{table*}

%% file: appendix/figures/ko_analysis_appendix.tex
\begin{figure*}[t!]
  \centering
  \includegraphics[width=0.9\textwidth]{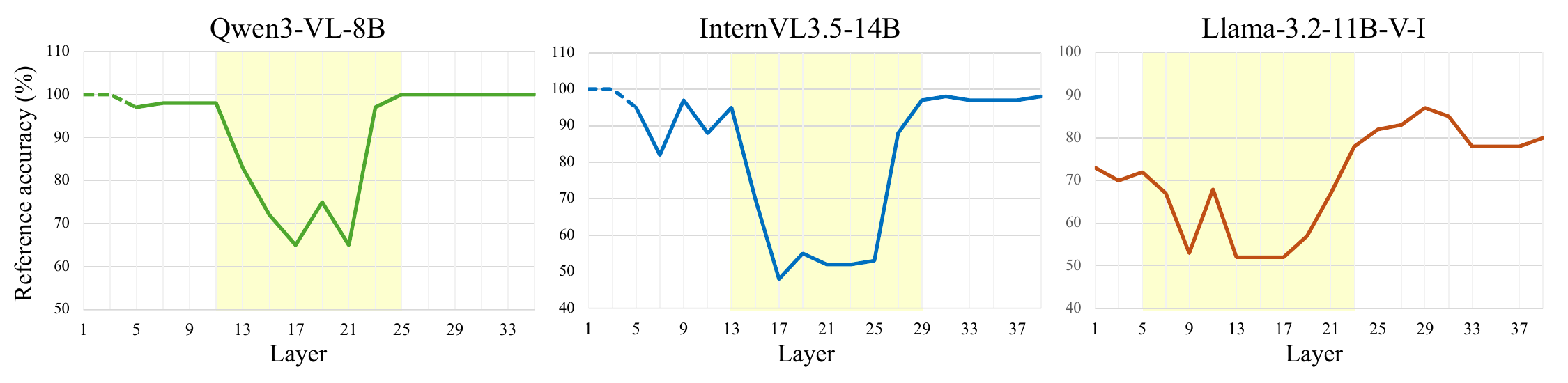}
  \vspace{-0.1cm}
  \caption{
    Layer-wise analysis of text-to-text attention blocking for Qwen3-VL-8B, InternVL3.5-14B, and Llama-3.2-11B-V-I on the captioning task.
    The yellow shaded region indicates the layer range selected by RSCD.
  }
  \vspace{-0.1cm}

  \label{fig:ko_analysis_appendix}
\end{figure*}

%% file: appendix/tables/knockout_text_only.tex
\begin{table}[t]
\centering
\small
\begin{tabular}{c|c|c}
\toprule
\textbf{Center Layer} & \textbf{EM} & \textbf{Sup-F1} \\
\midrule

1  & n/a   & n/a   \\
3  & 29.50 & n/a   \\
5  & 34.50 & 16.24 \\
7  & 35.00 & 12.39 \\
9  & 33.50 & 11.77 \\
11 & 26.50 & 8.62  \\
13 & 27.00 & \textbf{1.69} \\
15 & 23.00 & \textbf{1.79} \\
17 & 17.50 & \textbf{0.75} \\
19 & 27.50 & \textbf{2.78} \\
21 & 32.00 & 7.27  \\
23 & 25.00 & 15.46 \\
25 & 38.00 & 28.86 \\
27 & 34.50 & 27.08 \\

\bottomrule

\end{tabular}

\caption{Effect of layer-wise text-to-text attention masking on HotpotQA performance.}
\label{tab:text_only}
\vspace{-0.4cm}
\end{table}

%% file: appendix/tables/mvh_captioning_ko.tex
\begin{table}[t]
\centering
\small
\begin{tabular}{c|c}
\toprule
\textbf{Center Layer} & \textbf{Ref. Acc.}\\
\midrule
1  & n/a \\
3  & n/a \\
5  & 0.69 \\
7  & 0.70 \\
9  & 0.72 \\
11 & 0.74 \\
13 & \textbf{0.65} \\
15 & \textbf{0.51} \\
17 & \textbf{0.54} \\
19 & \textbf{0.59} \\
21 & 0.69 \\
23 & 0.69 \\
25 & 0.69 \\
27 & 0.69 \\
\bottomrule
\end{tabular}
\caption{Reference accuracy of LLaVA-OV-7B under layer-wise text-to-text attention blocking on the MVH-Bench validation subset.}
\label{tab:mvh_captioning_ko}
\end{table}

%% file: appendix/tables/rscd_generalization.tex
\begin{table*}[tb]

\centering
\footnotesize

\setlength{\tabcolsep}{2pt}
\renewcommand{\arraystretch}{1.25}

\begin{tabularx}{\textwidth}{@{}
>{\raggedright\arraybackslash}p{1.8cm}
!{\vrule width 0.9pt}
*{3}{>{\centering\arraybackslash}X}
!{\vrule width 0.5pt}
*{2}{>{\centering\arraybackslash}X}
!{\vrule width 0.5pt}
>{\centering\arraybackslash}X
!{\vrule width 0.9pt}
*{3}{>{\centering\arraybackslash}X}
!{\vrule width 0.5pt}
*{2}{>{\centering\arraybackslash}X}
!{\vrule width 0.5pt}
>{\centering\arraybackslash}X
!{\vrule width 0.9pt}
>{\centering\arraybackslash}X
@{}}

\toprule

\multirow{3}{*}{\textbf{LVLM}}
&
\multicolumn{6}{c!{\vrule width 0.9pt}}{\textbf{Cross-Instance}}
&
\multicolumn{6}{c!{\vrule width 0.9pt}}{\textbf{Cross-View}}
&
\multirow{3}{*}{\makecell{\textbf{MVH}\\\textbf{Score}}}
\\

\cmidrule(lr){2-7}
\cmidrule(lr){8-13}

&
\multicolumn{3}{c!{\vrule width 0.5pt}}{\textbf{Binary}}
&
\multicolumn{2}{c!{\vrule width 0.5pt}}{\textbf{M.C.}}
&
\multirow{2}{*}{\textbf{Score}}
&
\multicolumn{3}{c!{\vrule width 0.5pt}}{\textbf{Binary}}
&
\multicolumn{2}{c!{\vrule width 0.5pt}}{\textbf{M.C.}}
&
\multirow{2}{*}{\textbf{Score}}
&
\\

&
\textbf{Acc}
&
\textbf{p-Acc}
&
\textbf{q-Acc}
&
\textbf{Acc}
&
\textbf{p-Acc}
&
&
\textbf{Acc}
&
\textbf{p-Acc}
&
\textbf{q-Acc}
&
\textbf{Acc}
&
\textbf{p-Acc}
&
&
\\

\midrule
\multicolumn{14}{c}{\textbf{Qwen3-VL-8B}} \\

\midrule

Base model
& 71.46 & 53.19 & 28.33
& 75.69 & 59.44 & 288.12
& 70.76 & 52.92 & 29.17
& \underline{76.81} & \underline{58.33} & 287.99
& 576.11 \\

VCD
& 72.01 & 51.53 & 28.06
& \textbf{77.36} & \textbf{61.67} & \underline{290.62}
& \underline{72.50} & \textbf{55.28} & \underline{31.11}
& \textbf{78.33} & \textbf{59.44} & \textbf{296.67}
& \underline{587.29} \\

ICD
& \underline{72.36} & \underline{53.75} & \underline{31.11}
& 74.03 & 57.78 & 289.03
& 70.97 & 52.08 & 29.17
& 74.72 & 56.11 & 283.06
& 572.08 \\

\textbf{RSCD (Ours)}
& \textbf{73.47} & \textbf{56.39} & \textbf{33.33}
& \underline{76.11} & \underline{60.00} & \textbf{299.31}
& \textbf{73.68} & \underline{54.17} & \textbf{32.78}
& 75.69 & {57.78} & \underline{294.10}
& \textbf{593.40} \\

\midrule

\multicolumn{14}{c}{\textbf{InternVL3.5-14B}} \\

\midrule

Base model
& 62.36 & 34.17 & 12.78
& \textbf{75.69} & \textbf{58.89} & 243.89
& 70.28 & 47.08 & 25.56
& \textbf{80.83} & \underline{66.11} & 289.86
& 533.75 \\

VCD
& \underline{65.07} & \underline{39.31} & \underline{16.94}
& \underline{74.44} & \underline{56.67} & \underline{252.43}
& \underline{72.01} & \underline{51.39} & \underline{29.17}
& \textbf{80.83} & 65.00 & \underline{298.40}
& \underline{550.83} \\

ICD
& 62.99 & 34.72 & 14.72
& 74.17 & 55.83 & 242.43
& 68.06 & 44.72 & 20.56
& 80.00 & 64.17 & 277.50
& 519.93 \\

\textbf{RSCD (Ours)}
& \textbf{67.85} & \textbf{46.67} & \textbf{23.33}
& {73.89} & 55.00 & \textbf{266.74}
& \textbf{73.75} & \textbf{54.86} & \textbf{32.50}
& \underline{80.56} & \textbf{67.50} & \textbf{309.17}
& \textbf{575.90} \\

\midrule
\multicolumn{14}{c}{\textbf{Llama-3.2-11B-V-I}} \\

\midrule

Base model
& 54.72 & 20.83 & 5.00
& 56.39 & 29.44 & 166.39
& \textbf{50.83} & \underline{18.47} & \textbf{3.33}
& 45.28 & 13.89 & \underline{131.81}
& 298.19 \\

VCD
& \textbf{57.92} & \underline{24.17} & \underline{5.28}
& \underline{64.44} & 37.50 & \underline{189.31}
& \underline{50.76} & 17.08 & 0.83
& \underline{48.47} & 13.89 & 131.04
& \underline{320.35} \\

ICD
& 53.68 & 14.03 & 2.50
& 63.47 & \underline{38.33} & 172.01
& 50.14 & 15.97 & \underline{1.39}
& 47.64 & \textbf{14.72} & 129.86
& 301.88 \\

\textbf{RSCD (Ours)}
& \underline{57.85} & \textbf{31.53} & \textbf{7.22}
& \textbf{65.56} & \textbf{41.11} & \textbf{203.26}
& 49.65 & \textbf{20.69} & \underline{1.39}
& \textbf{49.44} & \underline{14.17} & \textbf{135.35}
& \textbf{338.61} \\

\bottomrule

\end{tabularx}

\caption{
Performance comparison of RSCD across different LVLM families and scales on MVH-Bench.
The best and second-best results are highlighted in \textbf{bold} and \underline{underline}, respectively.
}

\label{tab:rscd_generalization}

\end{table*}

%% file: appendix/tables/generalization_other_benchmark.tex
\begin{table}[t]
\centering
\small

\setlength{\tabcolsep}{4pt}
\renewcommand{\arraystretch}{1.15}

\begin{tabular}{@{}l|c|c@{}}
\toprule
\textbf{Benchmark} & \textbf{Method} & \textbf{Acc.} \\
\midrule

\multirow{2}{*}{MMBench}
& Baseline & 85.22 \\
& RSCD & \textbf{85.63} \\

\midrule

\multirow{2}{*}{SEED-Bench}
& Baseline & 68.50 \\
& RSCD & \textbf{70.00} \\

\bottomrule
\end{tabular}

\caption{
Evaluation of RSCD on MMBench and SEED-Bench.
}
\label{tab:generalization_mm}
\end{table}

\begin{table}[t]
\centering
\small
\setlength{\tabcolsep}{3pt}
\renewcommand{\arraystretch}{1.15}

\begin{tabular}{@{}l|c|c|c|c@{}}
\toprule
\textbf{Benchmark} & \textbf{Method}
& \textbf{Acc.}
& \textbf{F1}
& \textbf{Yes} \\
\midrule

\multirow{2}{*}{POPE}
& Baseline
& 87.38
& 86.52
& 43.62 \\

& RSCD
& \textbf{87.98}
& \textbf{86.79}
& 40.98 \\

\bottomrule
\end{tabular}

\caption{
Evaluation of RSCD on POPE.
}
\label{tab:generalization_pope}
\end{table}

\begin{table}[t]
\centering
\small
\setlength{\tabcolsep}{4pt}
\renewcommand{\arraystretch}{1.15}

\begin{tabular}{@{}l|c|c|c|c@{}}
\toprule
\textbf{Benchmark} & \textbf{Method}
& \textbf{CHAIRs}~$\downarrow$
& \textbf{CHAIRi}~$\downarrow$
& \textbf{Recall}~$\uparrow$ \\
\midrule

\multirow{2}{*}{CHAIR}
& Baseline
& \textbf{43.20}
& 11.07
& 65.15 \\

& RSCD
& 47.80
& \textbf{10.14}
& \textbf{67.74} \\

\bottomrule
\end{tabular}

\caption{
Evaluation of RSCD on CHAIR.
}
\label{tab:generalization_chair}
\end{table}

%% file: appendix/tables/max_view.tex
\begin{table*}[tb]

\centering
\footnotesize

\setlength{\tabcolsep}{2pt}
\renewcommand{\arraystretch}{1.3}

\begin{tabularx}{\textwidth}{@{}
>{\raggedright\arraybackslash}p{2.98cm}
!{\vrule width 0.9pt}
*{3}{>{\centering\arraybackslash}X}
!{\vrule width 0.5pt}
*{2}{>{\centering\arraybackslash}X}
!{\vrule width 0.5pt}
>{\centering\arraybackslash}X
!{\vrule width 0.9pt}
*{3}{>{\centering\arraybackslash}X}
!{\vrule width 0.5pt}
*{2}{>{\centering\arraybackslash}X}
!{\vrule width 0.5pt}
>{\centering\arraybackslash}X
!{\vrule width 0.9pt}
>{\centering\arraybackslash}X
@{}}

\toprule

\multirow{3}{*}{\textbf{Method}}
&
\multicolumn{6}{c!{\vrule width 0.9pt}}{\textbf{Cross-Instance}}
&
\multicolumn{6}{c!{\vrule width 0.9pt}}{\textbf{Cross-View}}
&
\multirow{3}{*}{\makecell{\textbf{MVH}\\\textbf{Score}}}
\\

\cmidrule(lr){2-7}
\cmidrule(lr){8-13}

&
\multicolumn{3}{c!{\vrule width 0.5pt}}{\textbf{Binary}}
&
\multicolumn{2}{c!{\vrule width 0.5pt}}{\textbf{M.C.}}
&
\multirow{2}{*}{\textbf{Score}}
&
\multicolumn{3}{c!{\vrule width 0.5pt}}{\textbf{Binary}}
&
\multicolumn{2}{c!{\vrule width 0.5pt}}{\textbf{M.C.}}
&
\multirow{2}{*}{\textbf{Score}}
&
\\

&
\textbf{Acc}
&
\textbf{p-Acc}
&
\textbf{q-Acc}
&
\textbf{Acc}
&
\textbf{p-Acc}
&
&
\textbf{Acc}
&
\textbf{p-Acc}
&
\textbf{q-Acc}
&
\textbf{Acc}
&
\textbf{p-Acc}
&
&
\\

\midrule

Base model (two-view)
& 54.22 & 27.26 & 0.00
& 64.17 & 35.83 & 181.48
& 50.00 & 23.75 & \underline{2.50}
& \textbf{60.00} & \textbf{27.50} & 163.75
& 345.23 \\

\midrule

Base model (max-view)
& 52.24 & 23.33 & 0.00
& 59.17 & 23.61 & 158.35
& 51.25 & 23.75 & 0.00
& 56.25 & \textbf{27.50} & 158.75
& 317.10 \\

VCD
& \underline{58.44} & 28.12 & 5.56
& \underline{65.56} & \underline{41.11} & \underline{198.79}
& 51.25 & 20.00 & \underline{2.50}
& \underline{58.75} & \textbf{27.50} & 160.00
& \underline{358.79} \\

ICD
& 57.95 & \underline{30.52} & \underline{5.62}
& 61.67 & 35.83 & 191.59
& 48.12 & 21.25 & 0.00
& 56.25 & 22.50 & 148.12
& 339.71 \\

AvisC
& 51.86 & 20.87 & 0.00
& 58.09 & 23.96 & 154.78
& \textbf{56.88} & \textbf{32.50} & \textbf{5.00}
& 53.75 & 17.50 & \underline{165.63}
& 320.41 \\

\textbf{RSCD (Ours)}
& \textbf{61.18} & \textbf{32.81} & \textbf{8.68}
& \textbf{66.94} & \textbf{41.39} & \textbf{210.99}
& \underline{55.00} & \underline{31.25} & 0.00
& \underline{58.75} & \underline{25.00} & \textbf{170.00}
& \textbf{380.99} \\

\bottomrule

\end{tabularx}

\caption{
Performance comparison on MVH-Bench under the original two-view and max-view settings.
The best and second-best results are highlighted in \textbf{bold} and \underline{underline}, respectively.
}

\label{tab:max_view}

\end{table*}

%% file: appendix/tables/free_form_captioning.tex
\begin{table}[t]
\centering
\small
\begin{tabular}{lcc}
\toprule
\textbf{Method} & \textbf{Ref. Acc.} & \textbf{Avg. \# Words} \\
\midrule
Base model & 69 & 10.01 \\
\textbf{RSCD (Ours)} & \textbf{77} & \textbf{10.50} \\
\bottomrule
\end{tabular}
\caption{Free-form captioning performance of RSCD on the MVH-Bench validation set.}
\label{tab:free_form_captioning}
\end{table}

%% file: appendix/tables/open_ended.tex
\begin{table}[t]
\centering
\small
\setlength{\tabcolsep}{3pt}
\renewcommand{\arraystretch}{1.15}

\begin{tabular}{@{}l|>{\centering\arraybackslash}p{2cm}|>{\centering\arraybackslash}p{2cm}|>{\centering\arraybackslash}p{1.35cm}@{}}
\toprule
\textbf{Method} & \textbf{\makecell{\textbf{Cross-Inst.}\\\textbf{Score}}}
& \textbf{\makecell{\textbf{Cross-View}\\\textbf{Score}}}
& \textbf{\makecell{\textbf{MVH}\\\textbf{Score}}}
 \\
\midrule

Base model
& 139.38
& 96.25
& 235.62 \\

\textbf{RSCD (Ours)}
& \textbf{153.12}
& \textbf{150.00}
& \textbf{303.12} \\

\bottomrule
\end{tabular}

\caption {Open-ended evaluation performance of RSCD on the MVH-Bench validation set.}
\label{tab:open_ended}
\end{table}

%% file: appendix/figures/statistics.tex
\begin{figure*}[t!]
  \centering
  \includegraphics[width=\textwidth]{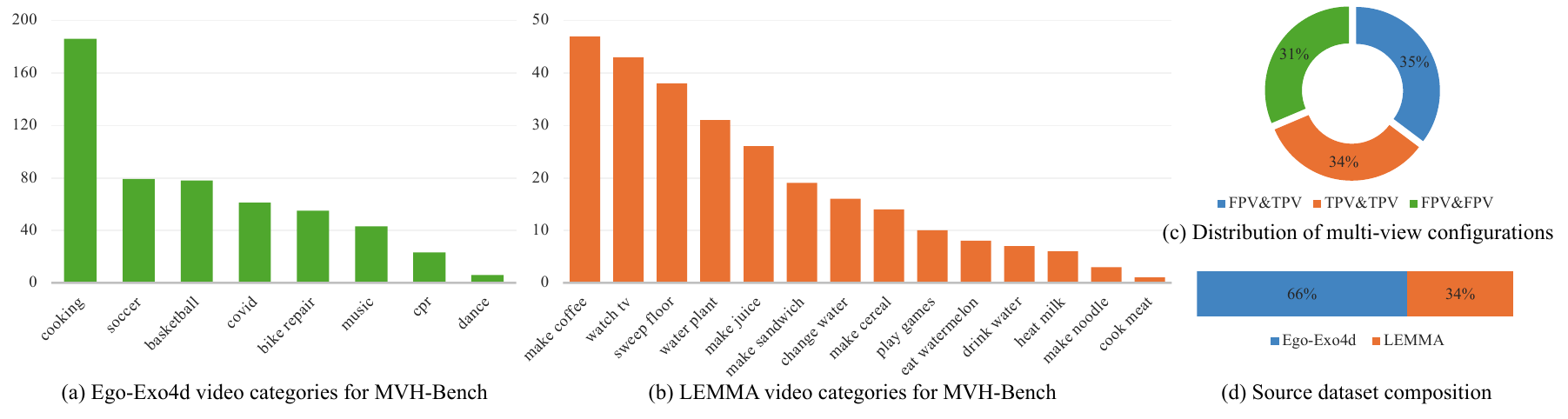}
  \caption{MVH-Bench statistics.
  Video category distribution of image pairs extracted from (a) Ego-Exo4D and  (b) LEMMA.
  (c) Distribution of image pair combinations across FPV and TPV.
  (d) Source dataset composition.
  }
  \label{fig:statistics}
\end{figure*}

%% file: appendix/figures/qualitative.tex
\begin{figure*}[t!]
  \centering
    \includegraphics[width=\linewidth]{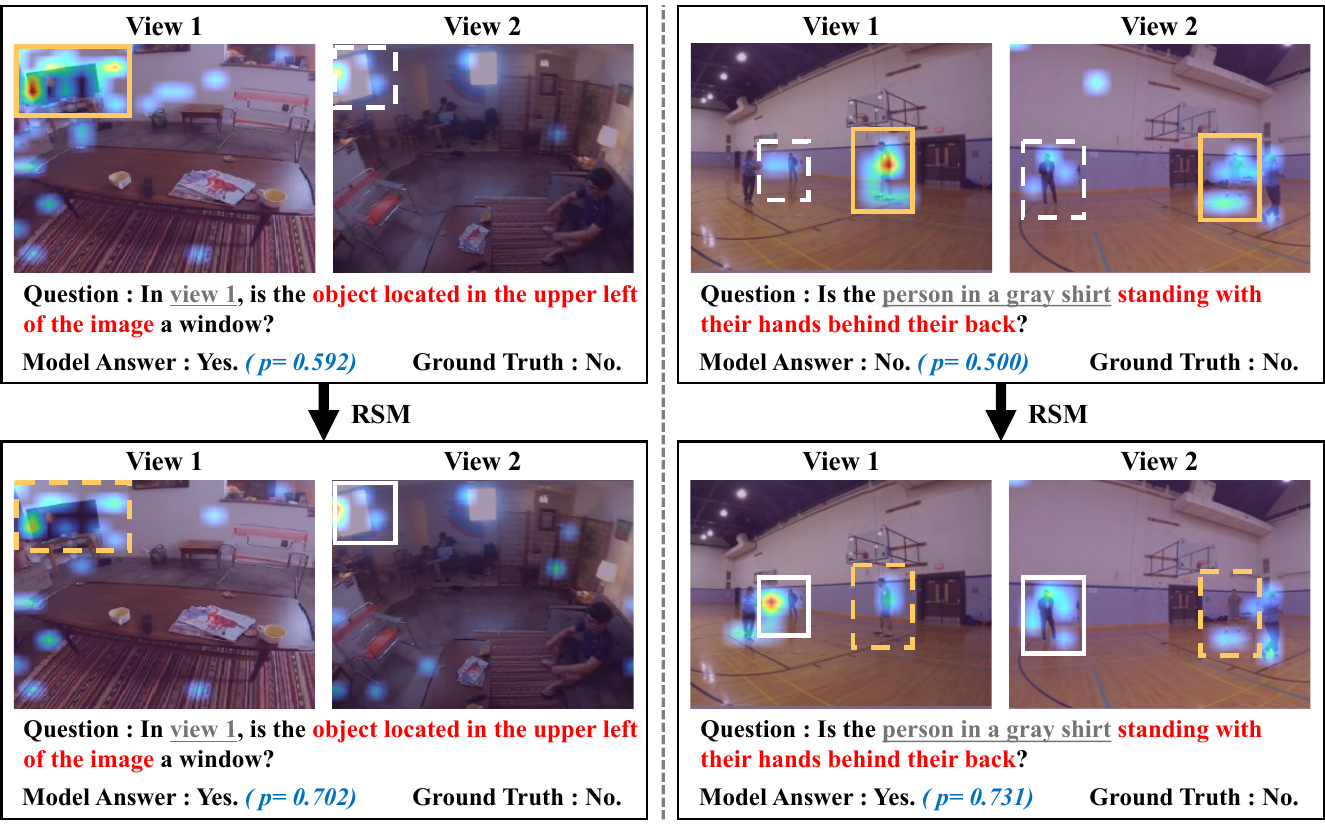}
  \caption{
Qualitative example illustrating the effect of reference shift masking.
Top: attention map of the base model.
Bottom: attention map with reference shift mask.
}
  \label{fig:qualitative_appendix}
\end{figure*}

%% file: appendix/figures/human_interface.tex
\begin{figure*}[t!]
  \centering
  \includegraphics[width=\textwidth]{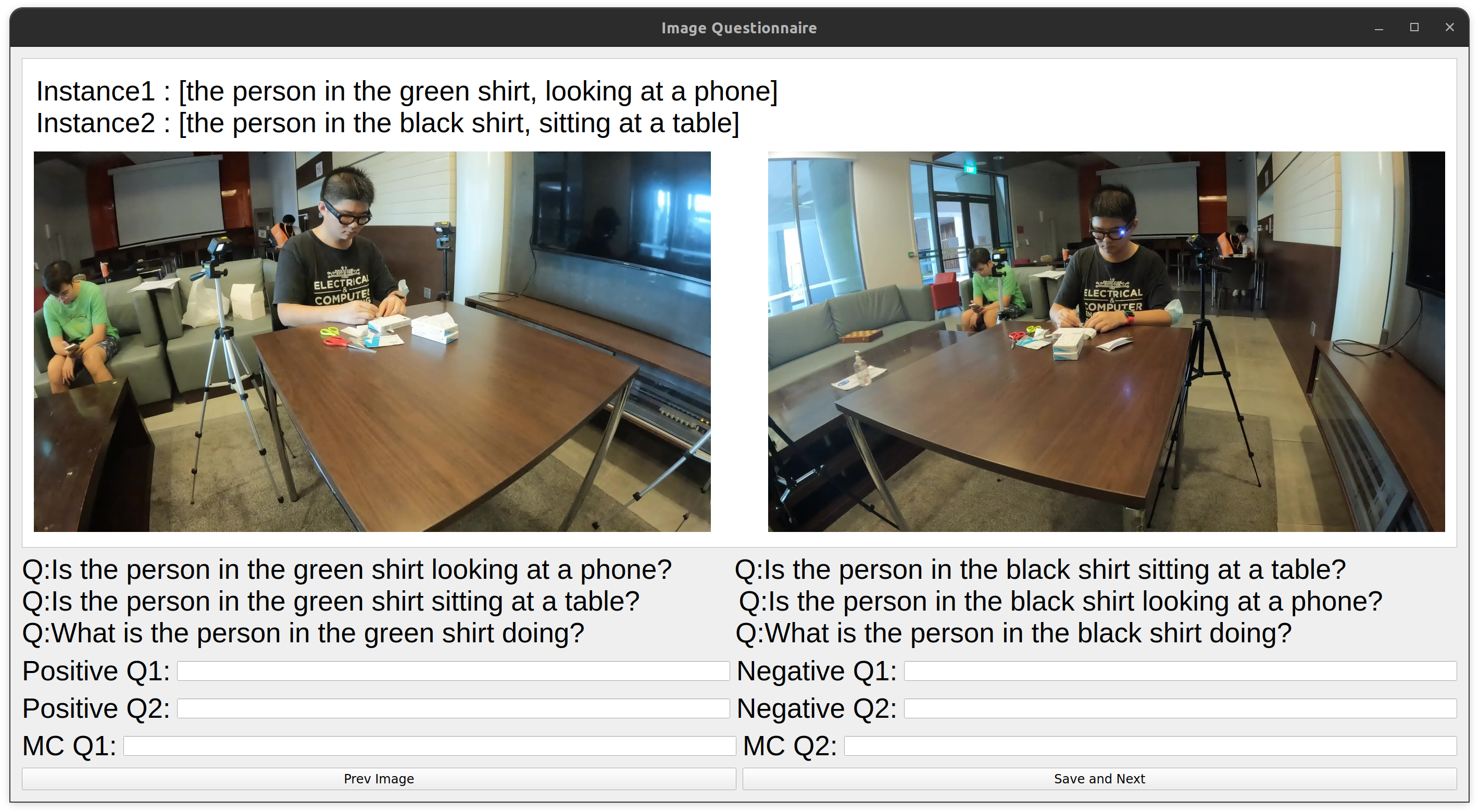}
  \caption{User interface used during the human verification process for constructing MVH-Bench.}
  \label{fig:interface}
\end{figure*}

%% file: appendix/tables/prompt.tex
\definecolor{myblue}{RGB}{230, 230, 230}
\tcbset{
  promptbox/.style={
    fonttitle=\scriptsize\bfseries,
    fontupper=\scriptsize\sffamily,
    enhanced,
    left=7.5pt,
    right=7.5pt,
    top=0pt,
    bottom=0pt,
    colback=white,
    colframe=black,
    colbacktitle=myblue,
    coltitle=black
  }
}

\lstdefinestyle{promptlisting}{
  basicstyle=\fontsize{6}{7}\selectfont\sffamily,
  breaklines=true,
  breakindent=0pt,
  escapeinside={(*@}{@*)}
}
\begin{figure*}[t]
\centering
\begin{tcolorbox}[
  promptbox,
  title={Instance-descriptor pair extraction prompt: action (first-person view image)},
]

\begin{lstlisting}[style=promptlisting]
(*@\textcolor{cyan}{\textbf{\{First-Person View Image\}}}@*)
You are given the visual input from the camera worn by the user (referred to as `I').

Based on this visual input, you must generate all person-action/posture description pairs that describe people and their actions or postures, strictly grounded in the visual content.

(*@\textbf{Instructions:}@*)
- Describe all people visible in the image using specific action or posture descriptions, grounded in visual evidence.
- Each pair must be a tuple: (`Person Description', `Action/Posture Description').
- Do not assign more than one action or posture to the same person; represent each individual with a single, most visually supported description.
- Avoid vague descriptions like `standing', `moving', `looking', or `background' unless they are detailed with context.
- Only generate up to five of the most clearly supported and valid descriptions based on the visual content.

(*@\textbf{Description Guidelines:}@*)
- `Person Description' should clearly refer to someone visible in the image, such as:
   `I'
   `The person wearing [color/item]'
   `The man standing near the sink'
- `Action/Posture Description' must be:
   In present participle (verb+ing) form for actions such as:
       `holding a cup'
       `typing on a keyboard'
   Or a posture-related phrase such as:
       `sitting cross-legged'
       `leaning back in a chair'

(*@\textbf{Output Format:}@*)
- Your final output must be a single Python list.
- Each item in the list must be a tuple:(`Person Description', `Action/Posture Description')

(*@\textbf{Validation Tips:}@*)
Use the following template questions to help verify the correctness of each tuple:
- Am/Is [Person Description] [Action Description]?
- Am/Is [Person Description] [Posture Description]?
If the answer is yes and it is visually supported, the tuple is likely valid.
Strictly follow the format shown in the examples below.

(*@\textbf{Examples:}@*)
[
  (`I', `sitting cross-legged'),
  (`the person wearing a green shirt', `raising one arm'),
  ...
]

\end{lstlisting}
\end{tcolorbox}
\caption{
Instance-descriptor pair extraction prompt: action (first-person view image).
}
\label{fig:action_first}
\end{figure*}

\begin{figure*}[t]
\begin{tcolorbox}[promptbox, title={Instance-descriptor pair extraction prompt: object (first-person view image)}]
\vspace{-0.5mm}
\begin{lstlisting}[style=promptlisting]
(*@\textcolor{cyan}{\textbf{\{First-Person View Image\}}}@*)
You are given the visual input from the camera worn by the user (referred to as `I').

Based on this visual input, you must generate all instance-attribute description pairs that describe instance along with their color, pattern, or object class, strictly grounded in the visual content.

(*@\textbf{Instructions:}@*)
- Identify all visible people and objects in the scene.
- For each, describe one relevant visual attribute: color, pattern, or object class.
- Each pair must be a tuple: (`Instance Description', `Attribute Description').
- Do not assign more than one attribute to the same instance; represent each object or person with a single, most visually supported description.
- Avoid vague descriptions like `person', `something', `object', or `background' unless they are detailed with context.
- Only generate up to five of the most clearly supported and valid descriptions based on the visual content.

(*@\textbf{Description Guidelines:}@*)
- `Instance Description' should clearly specify the instance (a visible person or object).
   `the shirt I am wearing'
   `the sweater worn by the person holding a cup'
   `the shoes I am wearing'
   `the object placed on the table'
   `the object I am holding in my right hand'

- `Attribute Description' must describe one of the following, based on what is visible:
   A color such as:
       `navy blue'
       `green'
   A pattern such as:
       `striped pattern'
       `checkered'
   An object class or category such as:
       `sneakers'
       `book'
       `mug cup'
       
(*@\textbf{Output Format:}@*)
- Your final output must be a single Python list.
- Each item in the list must be a tuple: (`Instance Description', `Attribute Description')

(*@\textbf{Validation Tips:}@*)
Use this question to check each pair:
- Is/Are [Instance Description] [Attribute Description]?
If the answer is yes and it is visually supported, the tuple is likely valid.
Strictly follow the format shown in the examples below.

(*@\textbf{Examples:}@*)
[
  (`the shirt I am wearing', `navy blue'),
  (`the shirt worn by the man wearing a cap', `white'),
  ...
]

\end{lstlisting}
\end{tcolorbox}
\caption{
Instance-descriptor pair extraction prompt: object (first-person view image).
}
\label{fig:object_first}
\end{figure*}

\definecolor{myblue}{RGB}{230, 230, 230}
\begin{figure*}[t]
\begin{tcolorbox}[promptbox, title={Instance-descriptor pair extraction prompt: numerical (first-person view image)}]
\vspace{-0.5mm}
\begin{lstlisting}[style=promptlisting]
(*@\textcolor{cyan}{\textbf{\{First-Person View Image\}}}@*)
You are given visual input from a camera worn by the user (referred to as `I').

Based on this visual input, you must generate all count-based description pairs that describe the number of instances or comparative counts of visible instances, strictly grounded in the visual content.

(*@\textbf{Instructions:}@*)
- Identify all visible people and objects that can be counted.
- For each case, generate outputs in the following formats:
   A count pair: (`Instance Description', `Number')
- Avoid vague descriptions such as `things', `items', `hand' `person', `object', or `background' unless they are detailed with context.
- All numbers must be integers from 1 to 5.
- Only generate up to five of the most clearly supported and valid descriptions based on the visual content.

(*@\textbf{Description Guidelines:}@*)
- `Instance Descriptions' should clearly specify the group of people or objects.
   `people near the window'
   `books on the shelf'
   `cups on the table'

(*@\textbf{Output Format:}@*)
- Your final output must be a single Python list.
- Each item in the list must be a tuple: (`Instance Description', `Number')

(*@\textbf{Validation Tips:}@*)
Use these questions to check each item:
- Is/Are there [Number] [Instance Description]?
If the answer is yes and it is visually supported, the entry is likely valid.
Strictly follow the format shown in the examples below.

(*@\textbf{Examples:}@*)
[
  (`people standing near the counter', 3),
  (`cups on the table', 2),
  ...
]

\end{lstlisting}
\end{tcolorbox}
\vspace{-1.0mm}
\caption{
Instance-descriptor pair extraction prompt: numerical (first-person view image).
}
\label{fig:numerical_first}
\vspace{4.0mm}
\end{figure*}

\definecolor{myblue}{RGB}{230, 230, 230}
\begin{figure*}[t]
\begin{tcolorbox}[promptbox, title={Instance-descriptor pair extraction prompt: spatial (first-person view image)}]
\vspace{-0.5mm}
\begin{lstlisting}[style=promptlisting]
(*@\textcolor{cyan}{\textbf{\{First-Person View Image\}}}@*)
You are given visual input from a camera worn by the user (referred to as `I').

Based on this visual input, you must generate all valid location-based description pairs that describe the location of an instance, strictly grounded in the visual content.

(*@\textbf{Instructions:}@*)
- Identify all visible people or objects whose positions are clearly identifiable, or whose relative positions can be meaningfully compared.
- For each case, generate outputs in the following formats:
   A location pair: (`Instance Description', `Location Description')
- Avoid vague descriptions such as `things', `items', `hand' `person', `object', or `background' unless they are detailed with context.
- Only generate up to five of the most clearly supported and valid descriptions based on the visual content.

(*@\textbf{Description Guidelines:}@*)
- `Instance Description' should be a clearly visible person or object in the scene, such as:
   `the blue mug cup'
   `the book on the shelf'
   `the person wearing [color/item]'

- `Location Description' refers to the absolute position of an instance, such as:
   `positioned above/below [object]'
   `to the left/right of [person]'
   `in front of [person]'
   `behind [person]'
   `in the top/bottom/left/right of the image'

(*@\textbf{Output Format:}@*)
- Your final output must be a single Python list.
- Each item in the list must be a tuple: (`Instance Description', `Location Description')

(*@\textbf{Validation Tips:}@*)
Use these questions to check each item:
- Is [Instance Description] [Location Description]?
If the answer is yes and it is visually supported, the entry is likely valid.
Strictly follow the format shown in the examples below.

(*@\textbf{Examples:}@*)
[
  (`the scissors', `in the right side of the image'),
  (`the tissue', `to my right'),
  ...
]

\end{lstlisting}
\end{tcolorbox}
\vspace{-1.0mm}
\caption{
Instance-descriptor pair extraction prompt: spatial (first-person view image).
}
\label{fig:spatial_first}
\end{figure*}

\definecolor{myblue}{RGB}{230, 230, 230}
\begin{figure*}[t]
\begin{tcolorbox}[promptbox, title={Instance-descriptor pair extraction prompt: action (third-person view image)}]
\vspace{-0.5mm}
\begin{lstlisting}[style=promptlisting]
(*@\textcolor{cyan}{\textbf{\{Third-Person View Image\}}}@*)
You are given the visual input from a fixed-position camera capturing a scene.

Based on this visual input, you must generate all person-action/posture description pairs that describe people and their actions or postures, strictly grounded in the visual content.

(*@\textbf{Instructions:}@*)
- Describe all people visible in the image using specific action or posture descriptions, grounded in visual evidence.
- Each pair must be a tuple: (`Person Description', `Action/Posture Description').
- Do not assign more than one action or posture to the same person; represent each individual with a single, most visually supported description.
- Avoid vague descriptions like `standing', `moving', `looking', or `background' unless they are detailed with context.
- Only generate up to five of the most clearly supported and valid descriptions based on the visual content.

(*@\textbf{Description Guidelines:}@*)
- `Person Description' should clearly refer to someone visible in the image, such as:
   `The person wearing [color/item]'
   `The man standing near the sink'
- `Action/Posture Description' must be:
   In present participle (verb+ing) form for actions such as:
       `holding a cup'
       `typing on a keyboard'
   Or a posture-related phrase such as:
       `sitting cross-legged'
       `leaning back in a chair'

(*@\textbf{Output Format:}@*)
- Your final output must be a single Python list.
- Each item in the list must be a tuple:(`Person Description', `Action/Posture Description')

(*@\textbf{Validation Tips:}@*)
Use the following template questions to help verify the correctness of each tuple:
- Is [Person Description] [Action Description]?
- Is [Person Description] [Posture Description]?
If the answer is yes and it is visually supported, the tuple is likely valid.
Strictly follow the format shown in the examples below.

(*@\textbf{Examples:}@*)
[
  (`the person wearing a watch', `sitting cross-legged'),
  (`the person wearing a green shirt', `raising one arm'),
  ...
]

\end{lstlisting}
\end{tcolorbox}
\caption{
Instance-descriptor pair extraction prompt: action (third-person view image).
}
\label{fig:action_third}
\end{figure*}

\definecolor{myblue}{RGB}{230, 230, 230}
\begin{figure*}[t]
\begin{tcolorbox}[promptbox, title={Instance-descriptor pair extraction prompt: object (third-person view image)}]
\vspace{-0.5mm}
\begin{lstlisting}[style=promptlisting]
(*@\textcolor{cyan}{\textbf{\{Third-Person View Image\}}}@*)
You are given the visual input from a fixed-position camera capturing a scene.

Based on this visual input, you must generate all instance-attribute description pairs that describe instances along with their color, pattern, or object class, strictly grounded in the visual content.

(*@\textbf{Instructions:}@*)
- Identify all visible people and objects in the scene.
- For each, describe one relevant visual attribute: color, pattern, or object class.
- Each pair must be a tuple: (`Instance Description', `Attribute Description').
- Do not assign more than one attribute to the same instance; represent each object or person with a single, most visually supported description.
- Avoid vague descriptions like `person', `something', `object', or `background' unless they are detailed with context.
- Only generate up to five of the most clearly supported and valid descriptions based on the visual content.

(*@\textbf{Description Guidelines:}@*)
- `Instance Description' should clearly specify the instance (a visible person or object).
   `the shirt worn by the person taking the photo'
   `the sweater worn by the person holding a cup'
   `the shoes worn by the woman walking on the path'
   `the object placed on the table'
   `the object held in the right hand of the person wearing a black watch'

- `Attribute Description' must describe one of the following, based on what is visible:
   A color such as:
       `navy blue'
       `green'
   A pattern such as:
       `striped pattern'
       `checkered'
   An object class or category such as:
       `sneakers'
       `book'
       `mug cup'
       
(*@\textbf{Output Format:}@*)
- Your final output must be a single Python list.
- Each item in the list must be a tuple: (`Instance Description', `Attribute Description')

(*@\textbf{Validation Tips:}@*)
Use this question to check each pair:
- Is/Are [Instance Description] [Attribute Description]?
If the answer is yes and it is visually supported, the tuple is likely valid.
Strictly follow the format shown in the examples below.

(*@\textbf{Examples:}@*)
[
  (`the shirt worn by the person taking the photo', `navy blue'),
  (`the shirt worn by the man wearing a cap', `white'),
  ...
]

\end{lstlisting}
\end{tcolorbox}
\caption{
Instance-descriptor pair extraction prompt: object (third-person view image).
}
\label{fig:object_third}
\end{figure*}

\definecolor{myblue}{RGB}{230, 230, 230}
\begin{figure*}[t]
\begin{tcolorbox}[promptbox, title={Instance-descriptor pair extraction prompt: numerical (third-person view image)}]
\vspace{-0.5mm}
\begin{lstlisting}[style=promptlisting]
(*@\textcolor{cyan}{\textbf{\{Third-Person View Image\}}}@*)
You are given the visual input from a fixed-position camera capturing a scene.

Based on this visual input, you must generate all count-based description pairs that describe the number of visible instances, strictly grounded in the visual content.

(*@\textbf{Instructions:}@*)
- Identify all visible people and objects that can be counted or compared.
- For each case, generate outputs in the following formats:
   A count pair: (`Instance Description', `Number')
- Avoid vague descriptions such as `things', `items', `hand', `person', `object', or `background' unless they are detailed with context.
- All numbers must be integers from 1 to 5.
- Only generate up to five of the most clearly supported and valid descriptions based on the visual content.

(*@\textbf{Description Guidelines:}@*)
- `Instance Descriptions' should clearly specify the group of people or objects.
   `people near the window'
   `books on the shelf'
   `cups on the table'

(*@\textbf{Output Format:}@*)
- Your final output must be a single Python list.
- Each item in the list must be a tuple: (`Instance Description', `Number')

(*@\textbf{Validation Tips:}@*)
Use these questions to check each item:
- Is/Are there [Number] [Instance Description]?
If the answer is yes and it is visually supported, the entry is likely valid.
Strictly follow the format shown in the examples below.

(*@\textbf{Examples:}@*)
[
  (`people standing near the counter', 3),
  (`cups on the table', 2),
  ...
]

\end{lstlisting}
\end{tcolorbox}
\vspace{-1.0mm}
\caption{
Instance-descriptor pair extraction prompt: numerical (third-person view image).
}
\label{fig:numerical_third}
\vspace{3.0mm}
\end{figure*}

\definecolor{myblue}{RGB}{230, 230, 230}
\begin{figure*}[t]
\begin{tcolorbox}[promptbox, title={Instance-descriptor pair extraction prompt: spatial (third-person view image)}]
\vspace{-0.5mm}
\begin{lstlisting}[style=promptlisting]
(*@\textcolor{cyan}{\textbf{\{Third-Person View Image\}}}@*)
You are given the visual input from a fixed-position camera capturing a scene.

Based on this visual input, you must generate all valid location-based description pairs that describe the location of an instance, strictly grounded in the visual content.

(*@\textbf{Instructions:}@*)
- Identify all visible people or objects whose positions are clearly identifiable, or whose relative positions can be meaningfully compared.
- For each case, generate outputs in the following formats:
   A location pair: (`Instance Description', `Location Description')
- Avoid vague descriptions such as `things', `items', `hand' `person', `object', or `background' unless they are detailed with context.
- Only generate up to five of the most clearly supported and valid descriptions based on the visual content.

(*@\textbf{Description Guidelines:}@*)
- `Instance Description' should be a clearly visible person or object in the scene, such as:
   `the blue mug cup'
   `the book on the shelf'
   `the person wearing [color/item]'

- `Location Description' refers to the absolute position of an instance, such as:
   `positioned above/below [object]'
   `to the left/right of [person]'
   `in front of [person]'
   `behind [person]'
   `in the top/bottom/left/right of the image'

(*@\textbf{Output Format:}@*)
- Your final output must be a single Python list.
- Each item in the list must be a tuple: (`Instance Description', `Location Description')

(*@\textbf{Validation Tips:}@*)
Use these questions to check each item:
- Is [Instance Description] [Location Description]?
If the answer is yes and it is visually supported, the entry is likely valid.
Strictly follow the format shown in the examples below.

(*@\textbf{Examples:}@*)
[
  (`the scissors', `in the right side of the image'),
  (`the tissue', `to the right of the person wearing a blue shirt'),
  ...
]

\end{lstlisting}
\end{tcolorbox}
\vspace{-1.0mm}
\caption{
Instance-descriptor pair extraction prompt: spatial (third-person view image).
}
\label{fig:spatial_third}
\end{figure*}